\documentclass[11pt]{article}

\usepackage[preprint]{acl}
\usepackage{times}
\usepackage{latexsym}
\usepackage[T1]{fontenc}
\usepackage[utf8]{inputenc}
\usepackage{microtype}
\usepackage{inconsolata}
\usepackage{graphicx}
\usepackage{booktabs,array,amssymb}
\usepackage{colortbl}
\usepackage{pifont}
\usepackage{multirow}
\usepackage{longtable}
\usepackage{newunicodechar}

\newunicodechar{ }{\nobreakspace{}}
\newunicodechar{–}{\textendash}
\newunicodechar{Ł}{\L{}}
\newunicodechar{ł}{\l{}}
\definecolor{MethodYesColor}{RGB}{0,160,0}
\definecolor{MethodNoColor}{RGB}{215,0,0}
\definecolor{MethodDateColor}{RGB}{78,99,130}
\newcommand{\MethodYes}{\textcolor{MethodYesColor}{\ding{51}}}
\newcommand{\MethodNo}{\textcolor{MethodNoColor}{\ding{55}}}
\newcommand{\MethodDate}[1]{{\fontsize{6.5}{7.3}\selectfont\textcolor{MethodDateColor}{\textit{(#1)}}}}
\DeclareMathSizes{7.8}{7}{5}{5}
\DeclareMathSizes{7.1}{7}{5}{5}
\DeclareMathSizes{7.2}{7}{5}{5}
\DeclareMathSizes{7.5}{7}{5}{5}
\DeclareMathSizes{6.5}{6}{5}{5}
\DeclareMathSizes{6.3}{6}{5}{5}

\graphicspath{{figures/}}

\title{A Task-Centric Survey of Long-Document Visual Question Answering:
Taxonomy, Evolution, and Future Directions}
\author{
  \textbf{Yufei Tong}$^{1}$ \quad \textbf{Guanjie Cheng}$^{2}$\thanks{\raggedright Corresponding authors: Guanjie Cheng (\href{mailto:chengguanjie@zju.edu.cn}{\texttt{chengguanjie@zju.edu.cn}}) and Leilei Zhang (\href{mailto:12460098@zju.edu.cn}{\texttt{12460098@zju.edu.cn}}).} \quad \textbf{Tianle Wu}$^{3}$ \quad \textbf{Leilei Zhang}$^{1}$\footnotemark[1] \\
  \textbf{Hang Zheng}$^{1,4}$ \quad \textbf{Peihan Wu}$^{1}$ \quad \textbf{Jianwei Yin}$^{1,2}$ \quad \textbf{Shuiguang Deng}$^{1}$ \\[3pt]
  \normalfont\normalsize $^{1}$College of Computer Science and Technology \\
  \normalfont\normalsize $^{2}$School of Software Technology \\
  \normalfont\normalsize $^{3}$Polytechnic Institute \\
  \normalfont\normalsize Zhejiang University, China \\
  \normalfont\normalsize $^{4}$BWTON Technology Co., Ltd., China
}

\newcommand{\ProjectLink}{\url{https://github.com/yufeiTongZJU/Long-Document-VQA-Survey}}

\hypersetup{
  pdftitle={A Task-Centric Survey of Long-Document Visual Question Answering: Taxonomy, Evolution, and Future Directions},
  pdfauthor={Yufei Tong; Guanjie Cheng; Tianle Wu; Leilei Zhang; Hang Zheng; Peihan Wu; Jianwei Yin; Shuiguang Deng},
  pdfsubject={A task-centric survey of long-document visual question answering},
  pdfkeywords={Document Understanding, Document VQA, Multimodal Retrieval-Augmented Generation, Information Retrieval, Agentic Reasoning}
}

\begin{document}
\maketitle

\begin{abstract}
Visual document understanding is essential for acquiring knowledge from visually rich documents and is attracting growing research interest.
Existing surveys emphasize architectures or technical approaches, with limited attention to how task bottlenecks shape technical choices and research priorities.
We present a task-centric survey that decomposes long-document visual question answering (VQA) into four components: document parsing, evidence organization and retrieval, generator organization and learning, and evaluation.
For each, we develop a fine-grained taxonomy and trace motivations, technical responses, and cross-component connections.
We focus on explicit retrieval while covering direct reading.
Our temporal analysis traces expanding priorities from multimodal content understanding to sufficient, high-fidelity evidence acquisition and sustained evidence management.
We outline future directions in evidence sufficiency, verifiable and correctable understanding, and budget-aware coordination, with an outlook on continual document understanding.\footnote{Project page: \ProjectLink.}
\end{abstract}

\par

\section{Introduction}
\label{sec:introduction}

Visually rich documents, such as research papers and financial reports, play a vital role in preserving and communicating knowledge through text, tables, charts, images, and their spatial and cross-modal relationships.
Visual document understanding (VDU) interprets this content and structure, encompassing foundational tasks such as information extraction, as well as higher-level visual question answering (VQA) \citep{chen2024svrag}.
While information extraction identifies target fields, entities, and relations, document VQA selects and integrates evidence to answer questions \citep{mathew2021docvqa}.

\begin{figure}[t]
  \centering
  \includegraphics[width=0.85\columnwidth]{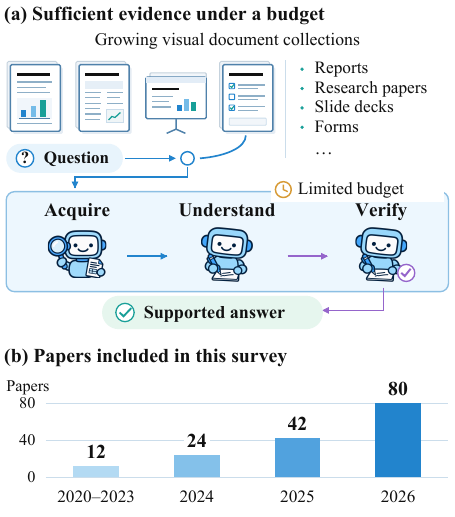}
  \caption{Core challenges of long-document VQA and publication trends of this field.}
  \label{fig:introduction}
  \vspace{-15pt}
\end{figure}

\begin{figure*}[t]
  \centering
  \includegraphics[width=0.95\textwidth]{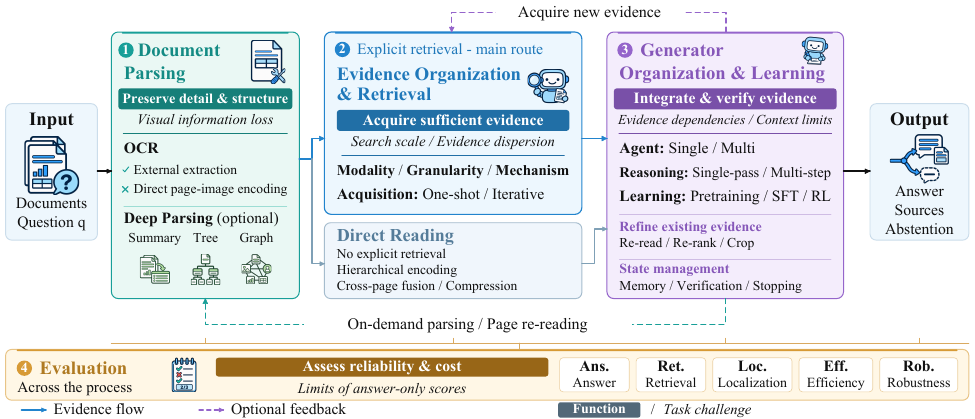}
  \caption{Task decomposition and two technical routes for long-document VQA. Colored bands indicate component functions, and italic labels identify the challenges they address. Both routes share parsing and generation, with evaluation throughout. Evidence sources and abstention are task-dependent outputs.}
  \label{fig:task-framework}
  \vspace{-10pt}
\end{figure*}

Document VQA has expanded from single-page understanding \citep{mathew2021docvqa} to evidence localization in MP-DocVQA \citep{tito2023hierarchical}, cross-page integration in SlideVQA \citep{tanaka2023slidevqa}, and long-context multimodal understanding in MMLongBench-Doc \citep{ma2024mmlongbench}.
Longer documents and larger collections increase the computational costs and context demands of exhaustive reading, while textualization or visual compression may discard essential chart, layout, and fine-grained textual information \citep{ref7,yu2025visrag}.
Evidence dispersed across pages, sections, or documents requires recovering relationships and integrating complementary information \citep{tanaka2023slidevqa,cho2025m3docvqa}.
We focus on long-document VQA: how systems can acquire, understand, and verify sufficient evidence to answer questions within a limited budget (Fig.~\ref{fig:introduction}).

Advances in multimodal large language models (MLLMs) and demand for complex document understanding are accelerating document VQA research.
Methods have expanded from multipage representations to visual retrieval-augmented generation (RAG) and feedback-driven evidence acquisition and reasoning \citep{tito2023hierarchical,yu2025visrag,wang2026vragrl}.
Rapid literature growth since 2024 (Fig.~\ref{fig:introduction}) motivates a systematic review of methods, benchmarks, and technical evolution.
Existing surveys cover document representations and multipage processing \citep{barboule2025survey}, MLLM representations and training \citep{ding2026survey}, multimodal retrieval \citep{gao2026scaling}, and visual retrievers and deployment \citep{yan2026unlocking}.
These surveys primarily organize the literature by model architecture and technical approach.
However, connections among long-document bottlenecks, fine-grained technical analysis, and evolving research motivations remain underexplored.

We analyze over 150 papers and organize this survey around task decomposition, fine-grained analysis, and temporal evolution.
Our contributions are threefold:
\textbf{(1) A unified, task-driven framework.} We map requirements for content fidelity, sufficient evidence acquisition, evidence use, and reliable evaluation onto four interconnected components: document parsing, evidence organization and retrieval, generator organization and learning, and evaluation.
\textbf{(2) A fine-grained technical taxonomy by task function.} We categorize representation and structure, retrieval modalities and granularities, evidence acquisition strategies, reasoning organization and learning objectives, and evaluation capabilities.
\textbf{(3) Technical evolution and future research directions.}
We trace how task bottlenecks shape technical choices as research priorities expand from multimodal understanding to evidence acquisition and sustained management.
We identify challenges and future directions in evidence sufficiency, verifiable and correctable understanding, and budget-aware coordination, with an outlook on continual document understanding.

\section{Task Formulation and Decomposition}
\label{sec:task-formulation}

\textbf{Task definition.}
Given a question $q$ and a candidate set of visual documents $\mathcal{D}$, long-document VQA produces an answer under a resource budget $B$:
\begin{equation}
  (\hat{a},\hat{\mathcal{S}})=F_{\theta}(q,\mathcal{D};B),
  \label{eq:task-formulation}
\end{equation}
where $F_{\theta}$ is the QA system, $\hat{a}$ its predicted answer, $\hat{\mathcal{S}}$ an optional set of evidence sources, and $B$ the resource limits on context, computation, and tool calls.
$\mathcal{D}$ contains one or more documents, each with ordered pages combining text, tables, charts, images, and layout information.
In the \emph{closed-domain} setting, the input document is specified for each question; in the \emph{open-domain} setting, systems identify relevant sources from a shared document collection \citep{yu2025visrag,cho2025m3docvqa,gao2026scaling}.
Pages preserve context and provenance, elements provide finer-grained evidence, and cross-page or cross-document relations link evidence needed jointly.

\begin{table*}[!t]
\centering
\captionsetup{position=top,skip=3pt}
\caption{Representative explicit-retrieval methods. (YY.MM): first release. Within task groups, methods follow final conference submission deadlines, then first release within each conference; preprints/journal papers are inserted by first release. Venue omits tracks. Full details: Appendix~\ref{app:explicit-retrieval}, Table~\ref{tab:explicit-methods-full}.}
\label{tab:explicit-methods}
\begingroup
\fontsize{7.8}{8.8}\selectfont
\setlength{\tabcolsep}{1.3pt}
\renewcommand{\arraystretch}{0.90}
\setlength{\extrarowheight}{0pt}
\setlength{\aboverulesep}{1.0pt}
\setlength{\belowrulesep}{1.0pt}
\resizebox{\textwidth}{!}{%
\begin{tabular}{@{}lccccclcccc@{}}
\toprule
\multicolumn{2}{c}{\textbf{Method}} & \multicolumn{2}{c}{\textbf{Parsing}} & \multicolumn{4}{c}{\textbf{Evidence Organization and Retrieval}} & \multicolumn{3}{c}{\textbf{Generation \& Learning}} \\
\cmidrule(lr){1-2}\cmidrule(lr){3-4}\cmidrule(lr){5-8}\cmidrule(l){9-11}
\textbf{Name} & \textbf{Venue} & \textbf{OCR} & \textbf{Deep} & \textbf{Modality} & \textbf{Gran.} & \textbf{Mechanism} & \textbf{Acquisition} & \textbf{Agent} & \textbf{Reasoning} & \textbf{Learning} \\
\midrule
\multicolumn{11}{@{}l@{}}{\textit{Retrieval-only methods}} \\
\addlinespace[0.8pt]
\textbf{DSE}~\citeyearpar{ma2024unifying}~\MethodDate{24.06} & EMNLP'24 & \MethodNo{} & \MethodNo{} & Uni(V) & P & Visual\textsuperscript{t} & One-shot & -- & -- & -- \\
\textbf{ColPali}~\citeyearpar{ref7}~\MethodDate{24.06} & ICLR'25 & \MethodNo{} & \MethodNo{} & Uni(V) & P & Visual\textsuperscript{t} & One-shot & -- & -- & -- \\
\textbf{LMS-Retrieval}~\citeyearpar{ref127}~\MethodDate{26.08} & ICDAR'26 & \MethodYes{} & Tree & Uni(T) & E & Text\textsuperscript{t} & One-shot & -- & -- & -- \\
\textbf{CMDR-Embed}~\citeyearpar{ref79}~\MethodDate{26.07} & ECCV'26 & \MethodNo{} & \MethodNo{} & Uni(V) & P & Visual\textsuperscript{t} & One-shot & -- & -- & -- \\
\textbf{MMRetHeads}~\citeyearpar{ref126}~\MethodDate{26.05} & EMNLP'26 & \MethodNo{} & \MethodNo{} & Uni(V) & P, E & Visual & One-shot & -- & -- & -- \\
\midrule
\multicolumn{11}{@{}l@{}}{\textit{Document VQA methods}} \\
\addlinespace[0.8pt]
\textbf{MHST}~\citeyearpar{ref86}~\MethodDate{22.07} & MM'22 & \MethodYes{} & \MethodNo{} & Uni(T) & P, E & Text & One-shot & Single & Single-pass & SFT \\
\textbf{CREAM}~\citeyearpar{ref118}~\MethodDate{24.10} & MM'24 & \MethodYes{} & \MethodNo{} & Uni(T) & P, E & Text\,+\,LLM & One-shot & Single & Single-pass & SFT \\
\textbf{VisRAG}~\citeyearpar{yu2025visrag}~\MethodDate{24.10} & ICLR'25 & \MethodNo{} & \MethodNo{} & Uni(V) & P & Visual\textsuperscript{t} & One-shot & Single & Single-pass & \MethodNo{} \\
\textbf{SV-RAG}~\citeyearpar{chen2024svrag}~\MethodDate{24.11} & ICLR'25 & \MethodNo{} & \MethodNo{} & Uni(V) & P & Visual\textsuperscript{t} & One-shot & Single & Single-pass & SFT \\
\textbf{VisDoMRAG}~\citeyearpar{ref10}~\MethodDate{24.12} & NAACL'25 & \MethodYes{} & \MethodNo{} & Mixed(T/V) & P, E & Text\,+\,Visual & One-shot & Multi & Single-pass & \MethodNo{} \\
\textbf{VDocRAG}~\citeyearpar{ref11}~\MethodDate{25.04} & CVPR'25 & \MethodNo{} & \MethodNo{} & Uni(V) & P & Visual\textsuperscript{t} & One-shot & Single & Single-pass & SFT \\
\textbf{Doc-React}~\citeyearpar{ref30}~\MethodDate{24.12} & ACL'25 & \MethodNo{} & \MethodNo{} & Uni(V) & P & Visual\,+\,VLM & Iterative & Single & Multi-step & \MethodNo{} \\
\textbf{NeuSym-RAG}~\citeyearpar{ref156}~\MethodDate{25.05} & ACL'25 & \MethodYes{} & Graph, Summary & Mixed(T/V) & D, P, E & Text\,+\,MM\,+\,VLM & Iterative & Single & Multi-step & \MethodNo{} \\
\textbf{MDocAgent}~\citeyearpar{ref32}~\MethodDate{25.03} & arXiv'25 & \MethodYes{} & \MethodNo{} & Mixed(T/V) & P & Text\,+\,Visual & One-shot & Multi & Single-pass & \MethodNo{} \\
\textbf{VRAG-RL}~\citeyearpar{wang2026vragrl}~\MethodDate{25.05} & NIPS'25 & \MethodNo{} & \MethodNo{} & Uni(V) & P, E & Visual\,+\,VLM\textsuperscript{t} & Iterative & Single & Multi-step & SFT+RL \\
\textbf{RA-Tuning}~\citeyearpar{ref16}~\MethodDate{24.11} & EMNLP'25 & \MethodYes{} & \MethodNo{} & Uni(V) & P, E & Visual & One-shot & Single & Single-pass & SFT \\
\textbf{ViDoRAG}~\citeyearpar{ref18}~\MethodDate{25.02} & EMNLP'25 & \MethodYes{} & \MethodNo{} & Mixed(T/V) & P & Text\,+\,Visual\,+\,VLM & One-shot & Multi & Multi-step & \MethodNo{} \\
\textbf{SimpleDoc}~\citeyearpar{ref34}~\MethodDate{25.06} & EMNLP'25 & \MethodYes{} & Summary & Mixed(T/V) & P & Visual\,+\,LLM & Iterative & Single & Multi-step & \MethodNo{} \\
\textbf{MoLoRAG}~\citeyearpar{ref35}~\MethodDate{25.09} & EMNLP'25 & \MethodNo{} & Graph & Uni(V) & P & Visual\,+\,VLM & One-shot & Single & Single-pass & \MethodNo{} \\
\textbf{LILaC}~\citeyearpar{ref36}~\MethodDate{25.11} & EMNLP'25 & \MethodYes{} & Graph & Mixed(T/V) & E & MM & One-shot & Single & Single-pass & \MethodNo{} \\
\textbf{DocAgent}~\citeyearpar{ref37}~\MethodDate{25.11} & EMNLP'25 & \MethodYes{} & Tree & Uni(T) & P, E & Text\,+\,VLM & Iterative & Multi & Multi-step & \MethodNo{} \\
\textbf{M3DocRAG}~\citeyearpar{cho2025m3docvqa}~\MethodDate{24.11} & ICCV'25 & \MethodNo{} & \MethodNo{} & Uni(V) & P & Visual & One-shot & Single & Single-pass & \MethodNo{} \\
\textbf{RegionRAG}~\citeyearpar{ref101}~\MethodDate{25.10} & AAAI'26 & \MethodNo{} & \MethodNo{} & Uni(V) & E & Visual\textsuperscript{t} & One-shot & Single & Single-pass & \MethodNo{} \\
\textbf{URaG}~\citeyearpar{ref38}~\MethodDate{25.11} & AAAI'26 & \MethodNo{} & \MethodNo{} & Uni(V) & P & Visual\textsuperscript{t} & One-shot & Single & Single-pass & SFT \\
\textbf{EXTRACTOR}~\citeyearpar{ref157}~\MethodDate{25.09} & ICLR'26 & \MethodYes{} & Graph & Uni(T) & D, P, E & Text\,+\,LLM\textsuperscript{t} & Iterative & Single & Multi-step & SFT \\
\textbf{SCoPE VLM}~\citeyearpar{ref70}~\MethodDate{25.10} & EACL'26 & \MethodNo{} & \MethodNo{} & Uni(V) & P & VLM\textsuperscript{t} & Iterative & Single & Multi-step & SFT+RL \\
\textbf{Doc-Researcher}~\citeyearpar{dong2026docresearcher}~\MethodDate{25.10} & WWW'26 & \MethodYes{} & Summary & Mixed(T/V) & D, P, E & MM\,+\,LLM & Iterative & Multi & Multi-step & \MethodNo{} \\
\textbf{DMAP}~\citeyearpar{ref40}~\MethodDate{26.01} & WWW'26 & \MethodYes{} & Tree, Summary & Mixed(T/V) & P, E & Text\,+\,Visual\,+\,LLM & Iterative & Multi & Multi-step & \MethodNo{} \\
\textbf{RAG-Anything}~\citeyearpar{ref42}~\MethodDate{25.10} & arXiv'25 & \MethodYes{} & Graph & Uni(T) & E & Text & One-shot & Single & Single-pass & \MethodNo{} \\
\textbf{DocPrune}~\citeyearpar{ref44}~\MethodDate{26.04} & CVPR'26 & \MethodNo{} & \MethodNo{} & Uni(V) & P & Visual & One-shot & Single & Single-pass & \MethodNo{} \\
\textbf{CogDoc}~\citeyearpar{ref46}~\MethodDate{25.12} & arXiv'25 & \MethodNo{} & \MethodNo{} & Uni(V) & P & VLM\textsuperscript{t} & One-shot & Single & Single-pass & RL \\
\textbf{DocLens}~\citeyearpar{ref48}~\MethodDate{25.11} & ACL'26 & \MethodYes{} & \MethodNo{} & Mixed(T/V) & P, E & VLM & One-shot & Multi & Single-pass & \MethodNo{} \\
\textbf{Doc-V*}~\citeyearpar{ref50}~\MethodDate{26.04} & ACL'26 & \MethodNo{} & \MethodNo{} & Uni(V) & P & Visual\,+\,VLM\textsuperscript{t} & Iterative & Single & Multi-step & SFT+RL \\
\textbf{MM-Doc-R1}~\citeyearpar{ref51}~\MethodDate{26.04} & ACL'26 & \MethodYes{} & Tree & Uni(T) & P & Text & Iterative & Multi & Multi-step & RL \\
\textbf{VimRAG}~\citeyearpar{ref53}~\MethodDate{26.02} & ICML'26 & \MethodNo{} & \MethodNo{} & Uni(V) & P & MM\,+\,VLM\textsuperscript{t} & Iterative & Single & Multi-step & SFT+RL \\
\textbf{MoDora}~\citeyearpar{ref94}~\MethodDate{26.02} & PACMMOD'26 & \MethodYes{} & Tree, Summary & Mixed(T/V) & E & Text\,+\,LLM\,+\,VLM & One-shot & Single & Single-pass & \MethodNo{} \\
\textbf{VisRAG 2.0}~\citeyearpar{ref41}~\MethodDate{25.10} & MM'26 & \MethodNo{} & \MethodNo{} & Uni(V) & P & Visual & One-shot & Single & Single-pass & SFT+RL \\
\textbf{AgenticOCR}~\citeyearpar{ref104}~\MethodDate{26.02} & NIPS'26 & \MethodYes{} & \MethodNo{} & Mixed(T/V) & P, E & VLM\textsuperscript{t} & One-shot & Single & Multi-step & SFT+RL \\
\textbf{DocAtlas}~\citeyearpar{ref72}~\MethodDate{26.07} & NIPS'26 & \MethodYes{} & Tree, Summary & Mixed(T/V) & P, E & LLM\,+\,VLM\textsuperscript{t} & Iterative & Single & Multi-step & RL \\
\textbf{DocDancer}~\citeyearpar{ref52}~\MethodDate{26.01} & EMNLP'26 & \MethodYes{} & Tree & Uni(T) & P, E & Text & Iterative & Single & Multi-step & SFT \\
\textbf{AWM}~\citeyearpar{ref96}~\MethodDate{26.08} & EMNLP'26 & \MethodNo{} & \MethodNo{} & Uni(V) & P & MM & Iterative & Single & Multi-step & SFT+RL \\
\textbf{Doc-REFRAG}~\citeyearpar{ref59}~\MethodDate{26.08} & EMNLP'26 & \MethodNo{} & \MethodNo{} & Uni(V) & P & MM & One-shot & Single & Single-pass & SFT+RL \\
\textbf{EviProp}~\citeyearpar{ref115}~\MethodDate{26.06} & arXiv'26 & \MethodYes{} & Graph & Mixed(T/V) & P & Visual\,+\,Text & One-shot & Single & Single-pass & \MethodNo{} \\
\textbf{MAGE-RAG}~\citeyearpar{ref56}~\MethodDate{26.06} & arXiv'26 & \MethodYes{} & Graph, Summary & Mixed(T/V) & P, E & Visual\,+\,VLM & Iterative & Single & Multi-step & \MethodNo{} \\
\textbf{HierDoc}~\citeyearpar{ref134}~\MethodDate{26.07} & arXiv'26 & \MethodYes{} & \MethodNo{} & Mixed(T/V) & P, E & VLM\textsuperscript{t} & One-shot & Single & Single-pass & \MethodNo{} \\
\textbf{DocTrace}~\citeyearpar{ref60}~\MethodDate{26.08} & arXiv'26 & \MethodYes{} & \MethodNo{} & Uni(V) & P, E & VLM\textsuperscript{t} & One-shot & Single & Single-pass & SFT+RL \\
\textbf{DocMemo}~\citeyearpar{ref148}~\MethodDate{26.08} & arXiv'26 & \MethodYes{} & Tree, Summary & Mixed(T/V) & P & Visual\,+\,LLM & Iterative & Multi & Multi-step & \MethodNo{} \\
\textbf{D2-ScaleAgent}~\citeyearpar{ref149}~\MethodDate{26.08} & arXiv'26 & \MethodNo{} & \MethodNo{} & Uni(V) & P, E & Visual\,+\,VLM & Iterative & Multi & Multi-step & \MethodNo{} \\
\textbf{DocClaw}~\citeyearpar{ref145}~\MethodDate{26.08} & arXiv'26 & \MethodYes{} & Tree & Mixed(T/V) & P, E & Text\,+\,Visual\,+\,VLM & Iterative & Single & Multi-step & \MethodNo{} \\
\textbf{SCORE}~\citeyearpar{ref150}~\MethodDate{26.09} & arXiv'26 & \MethodNo{} & \MethodNo{} & Uni(V) & P, E & Visual\,+\,VLM\textsuperscript{t} & Iterative & Single & Multi-step & SFT+RL \\
\bottomrule
\end{tabular}%
}
\par\vspace{1.5pt}\nointerlineskip
\begin{minipage}{\linewidth}
\fontsize{7.5}{8.5}\selectfont \textbf{Notation:} \MethodNo{} = absent; D/P/E = Document/Page/Element. Mechanism groups retriever families; $+$ combines families; $t$ marks added retrieval, scoring, or page-selection training for at least one component in the marked family. \textbf{Retriever families:} Visual = visual page/region retrieval; Text = text retrieval/reranking; MM = multimodal retrieval; LLM/VLM = language/vision-language models for selection, reranking, navigation, or region refinement.
\end{minipage}
\endgroup
\vspace{-15pt}
\end{table*}

Long documents pose three challenges.
\emph{(i) Input and search scale:} more candidate pages expand search spaces and increase context and computational demands for exhaustive reading.
\emph{(ii) Visual content complexity:} information in small text, charts, and spatial layouts may be lost through resizing, conversion to text, or compression.
\emph{(iii) Evidence distribution and dependencies:} answers may require complementary evidence across pages or documents; local relevance does not ensure sufficiency \citep{ma2024mmlongbench,yu2025visrag}.

\textbf{Four analytical components.}
Four interconnected components address these challenges.
\textbf{\emph{(1) Document parsing}} addresses visual information loss by preserving details and structure.
Content extraction, visual preservation, and semantic structure construction provide readable representations and navigable structures, balancing fidelity against processing cost.
\textbf{\emph{(2) Evidence organization and retrieval}} addresses search scale and dispersion by acquiring sufficient evidence amid noise.
It locates and selects evidence using document representations and structures, passing it and its sources to the generator.
\textbf{\emph{(3) Generator organization and learning}} addresses evidence dependencies and context limits through evidence integration and verification.
It covers evidence presentation, organizational forms, reasoning processes, and learning objectives under computational constraints; intermediate judgments can guide acquisition, rereading, or verification.
\textbf{\emph{(4) Evaluation}} goes beyond answer scores to assess correctness, grounding, efficiency, and reliability throughout the process.
Checks of evidence acquisition and use or controlled experiments diagnose parsing, retrieval, and reasoning errors and compare techniques' benefits and costs.

\textbf{Two technical routes.}
Fig.~\ref{fig:task-framework} connects these components through \textbf{\emph{direct reading}} and \textbf{\emph{explicit retrieval}}.
Direct reading answers from given document representations and may use hierarchical encoding, cross-page fusion, or compression to manage long-input costs \citep{tito2023hierarchical}.
Explicit retrieval selects question-relevant documents, pages, or elements (e.g., text blocks, tables, and region crops) for the generator to read \citep{yu2025visrag,cho2025m3docvqa,dong2026docresearcher}.
Reasoning feedback can guide further evidence acquisition \citep{wang2026vragrl}, local rereading, or on-demand parsing of existing evidence.
These components describe task functions; their implementation and execution order may vary.
Explicit retrieval accounts for approximately 80\% of surveyed methods, motivating our focus on this route while also covering direct reading.
Sec.~\ref{sec:taxonomy} develops a fine-grained taxonomy by task function and traces each component's research motivations and technical evolution.

\section{Taxonomy and Technical Evolution}
\label{sec:taxonomy}
For each component, we develop a fine-grained taxonomy, compare technical benefits and costs, and trace evolving research motivations and technical responses.
Stages reflect expanding research priorities: earlier approaches coexist, and methods may adopt techniques from different stages across components.
Tables~\ref{tab:explicit-methods} and~\ref{tab:direct-methods} compare representative explicit-retrieval and direct-reading methods; full comparisons appear in Appendices~\ref{app:direct-reading} and~\ref{app:explicit-retrieval}.

\begin{table}[!t]
\centering
\captionsetup{position=top,skip=3pt}
\caption{Taxonomy of representative direct-reading methods. Full comparison: Appendix~\ref{app:direct-reading}, Table~\ref{tab:direct-methods-full}.}
\label{tab:direct-methods}
\begingroup
\fontsize{7.8}{8.8}\selectfont
\setlength{\tabcolsep}{1.3pt}
\renewcommand{\arraystretch}{0.90}
\setlength{\extrarowheight}{0pt}
\setlength{\aboverulesep}{1.0pt}
\setlength{\belowrulesep}{1.0pt}
\resizebox{\columnwidth}{!}{%
\begin{tabular}{@{}lcccc@{}}
\toprule
\textbf{Method} & \textbf{Venue} & \textbf{OCR} & \textbf{Model} & \textbf{Learning} \\
\midrule
\multicolumn{5}{@{}l@{}}{\textit{Single-page VQA}} \\
\addlinespace[0.8pt]
\textbf{Donut}~\citeyearpar{ref67}~\MethodDate{21.11} & ECCV'22 & \MethodNo{} & BART & PT+SFT \\
\textbf{LayoutLMv3}~\citeyearpar{huang2022layoutlmv3}~\MethodDate{22.04} & MM'22 & \MethodYes{} & RoBERTa & PT+SFT \\
\textbf{Pix2Struct}~\citeyearpar{ref76}~\MethodDate{22.10} & ICML'23 & \MethodNo{} & Pix2Struct & PT+SFT \\
\textbf{DocVLM*}~\citeyearpar{ref25}~\MethodDate{24.12} & CVPR'25 & \MethodYes{} & Qwen-VL & SFT \\
\midrule
\multicolumn{5}{@{}l@{}}{\textit{Multi-page VQA}} \\
\addlinespace[0.8pt]
\textbf{M3D}~\citeyearpar{tanaka2023slidevqa}~\MethodDate{23.01} & AAAI'23 & \MethodYes{} & T5 & SFT \\
\textbf{Hi-VT5}~\citeyearpar{tito2023hierarchical}~\MethodDate{22.12} & PR'23 & \MethodYes{} & T5 & PT+SFT \\
\textbf{GRAM}~\citeyearpar{ref117}~\MethodDate{24.01} & CVPR'24 & \MethodYes{} & T5 & SFT \\
\textbf{mPLUG-DocOwl2}~\citeyearpar{ref26}~\MethodDate{24.09} & ACL'25 & \MethodNo{} & Llama & PT+SFT \\
\textbf{DocR1}~\citeyearpar{ref27}~\MethodDate{25.08} & AAAI'26 & \MethodNo{} & Qwen-VL & RL \\
\textbf{DocSeeker}~\citeyearpar{ref29}~\MethodDate{26.04} & CVPR'26 & \MethodNo{} & Qwen-VL & SFT+RL \\
\bottomrule
\end{tabular}%
}
\par\vspace{1.5pt}\nointerlineskip
\begin{minipage}{\linewidth}
\fontsize{7.5}{8.5}\selectfont All methods use a Single agent and no Deep Parsing. PT = Pretraining. Model lists backbone families only. DocVLM* primarily evaluated on single pages but also supports multiple pages.
\end{minipage}
\endgroup
\vspace{-15pt}
\end{table}

\subsection{Document Parsing}
\label{sec:parsing}

\begin{figure}[t]
  \centering
  \includegraphics[width=0.9\columnwidth]{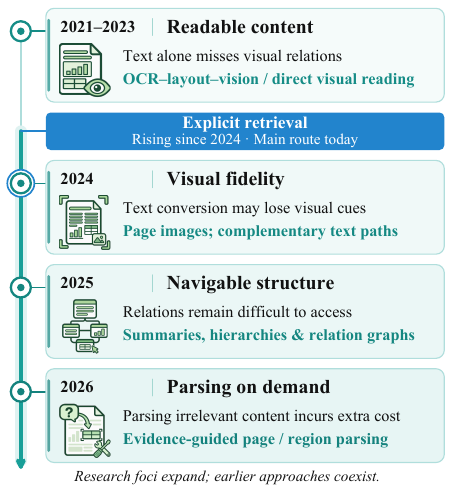}
  \caption{Four stages of document parsing: bottlenecks in black, technical responses in teal.}
  \label{fig:parsing-evolution}
  \vspace{-15pt}
\end{figure}

\textbf{Technical taxonomy.}
Document parsing converts raw pages into accessible content and structure.
Tables~\ref{tab:explicit-methods} and~\ref{tab:direct-methods} distinguish two independent dimensions: \emph{OCR} and \emph{Deep Parsing}.
OCR~(\MethodYes{}) denotes dependence on external extraction or textualization, including OCR, PDF text extraction, and VLM transcription or description of charts and tables.
Extracted text supports indexing and language processing, but linearization may lose layout, text--image correspondences, and table structure, while recognition errors can propagate downstream \citep{ma2024unifying,ref7,yu2025visrag,chen2024svrag}.
OCR~(\MethodNo{}) methods primarily encode pages visually to reduce textualization loss but remain limited by resolution and visual reading ability \citep{yu2025visrag,chen2024svrag}.
Methods combining extracted or transcribed text with page images are marked OCR~(\MethodYes{}) \citep{ref10,ref32}.
Deep Parsing~(\MethodYes{}) builds semantic overviews or explicit relations: \emph{Summary} captures page or document semantics, \emph{Tree} represents section hierarchies and containment, and \emph{Graph} organizes text--image, entity, or cross-page relations \citep{ref34,ref35,ref36,ref37}; these structures can be combined.
Element transcription or description alone does not count as Deep Parsing.
Parsing may cover full documents in advance or proceed on demand, guided by the question or ongoing process.

\textbf{Evolution of research focus (Fig.~\ref{fig:parsing-evolution}).}
Early work prioritized \emph{readable content} while preserving layout and visual relations.
LayoutLMv3 and UDOP jointly model OCR text, layout, and images \citep{huang2022layoutlmv3,ref87}, while Donut, Pix2Struct, and UReader learn directly from pages or screenshots \citep{ref67,ref76,ref92}.
From 2024, explicit retrieval emphasized \emph{original-page visual fidelity}: DSE, ColPali, VisRAG, and SV-RAG use page images directly \citep{ma2024unifying,ref7,yu2025visrag,chen2024svrag}, while VisDoMRAG and MDocAgent combine visual and textual paths to mitigate omissions and reading biases \citep{ref10,ref32}.
In 2025, \emph{navigable structure} made content relations accessible: SimpleDoc and Doc-Researcher summarize pages and documents, respectively \citep{ref34,dong2026docresearcher}; DocAgent builds section hierarchies \citep{ref37}; and MoLoRAG and LILaC construct page-similarity and layered component graphs, respectively \citep{ref35,ref36}.
To avoid parsing irrelevant content at scale, work in 2026 explores \emph{parsing on demand}: AgenticOCR performs question-guided regional OCR \citep{ref104}; DocLens, HierDoc, and DocTrace extract elements or parse selected pages \citep{ref48,ref134,ref60}; and DocClaw adapts parsing tool use to tasks and intermediate results \citep{ref145}.

Parsing supports evidence acquisition (Sec.~\ref{sec:retrieval}); its value should be assessed by evidence completeness, answer quality, and cost, rather than transcription accuracy or structural richness alone.

\subsection{Evidence Organization and Retrieval}
\label{sec:retrieval}

\textbf{Technical taxonomy.}
Evidence organization and retrieval select and assemble evidence from parsed content and structures, but local relevance does not ensure sufficient support for cross-page reasoning \citep{cho2025m3docvqa}.
Table~\ref{tab:explicit-methods} compares four dimensions.
\emph{Modality} specifies evidence-side matching representations: Uni(T) for text, Uni(V) for vision, and Mixed(T/V) for both.
\emph{Granularity} denotes evidence units supplied to generators or returned by retrieval-only methods: Document covers document-level evidence; Page covers full-page images or complete parsed page content; Element covers independently accessible text blocks, sections, charts, tables, or region crops. 
Pages preserve context but may add noise, whereas elements emphasize details but risk losing context.
Modality or granularity may change before generation, as when ViDoRAG maps text chunks to source page images \citep{ref18}; Table~\ref{tab:explicit-methods-full} details these mappings.
\emph{Mechanism} includes lexical or unimodal/multimodal vector matching and LLM/VLM-based page selection, reranking, or region selection, optionally combined with tree navigation or graph expansion.
The table lists retriever families.
\emph{Acquisition} distinguishes feedback-driven expansion: Iterative uses generation or control feedback to acquire new evidence beyond the current set; One-shot permits fixed multistage retrieval, reranking, or refinement without such expansion.

\begin{figure}[t]
  \centering
  \includegraphics[width=0.9\columnwidth]{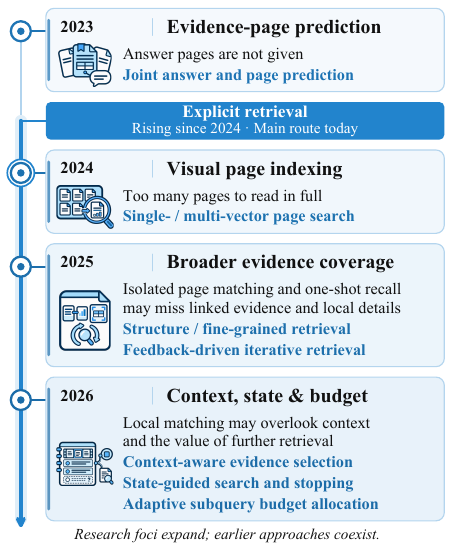}
  \caption{Four stages of evidence organization and retrieval: bottlenecks in black, technical responses in blue.}
  \label{fig:retrieval-evolution}
  \vspace{-15pt}
\end{figure}

\textbf{Evolution of research focus (Fig.~\ref{fig:retrieval-evolution}).}
Hi-VT5 and M3D jointly learn QA and evidence-page prediction over given multipage inputs \citep{tanaka2023slidevqa,tito2023hierarchical}, providing a precursor to explicit retrieval.
In 2024, input capacity limits motivated \emph{visual page indexing} and the rise of explicit retrieval.
DSE uses single-vector screenshot embeddings \citep{ma2024unifying}; ColPali uses multivector late interaction \citep{ref7}; and VisRAG, SV-RAG, and M3DocRAG connect page retrieval to visual QA \citep{yu2025visrag,chen2024svrag,cho2025m3docvqa}.
In 2025, incomplete coverage of linked evidence and local details motivated \emph{structural relations, fine-grained acquisition, and feedback-driven iterative retrieval}.
MoLoRAG and LILaC link complementary evidence through page or component relations \citep{ref35,ref36}; RegionRAG and Doc-Researcher support regional and multigranular acquisition, respectively \citep{ref101,dong2026docresearcher}; and VRAG-RL updates queries, retrieves additional pages, and refines acquired image regions during reasoning \citep{wang2026vragrl}.
In 2026, attention expands to \emph{context, explicit evidence states, and budget constraints}.
Doc-V* uses document-wide thumbnails for page navigation \citep{ref50}, while CMDR-Embed incorporates multipage context into page representations \citep{ref79}.
MAGE-RAG tracks page and element states \citep{ref56}; DocAtlas updates retrieval structures with reading-derived evidence \citep{ref72}; and DocMemo updates page beliefs and question memory to guide search, reading, and stopping \citep{ref148}.
MAB-DQA allocates retrieval budgets by estimating each subquestion's acquisition value from VLM feedback \citep{ref49}.

More evidence may add noise and redundancy without ensuring sufficient support.
Sec.~\ref{sec:generation} examines how generators integrate and verify evidence and guide further acquisition through feedback.

\subsection{Generation and Learning}
\label{sec:generation}
\textbf{Technical taxonomy.}
Generation and learning produce grounded answers under context and computational budgets.
Reading all pages preserves context at increasing input cost, whereas reading retrieved evidence narrows inputs but must address evidence gaps and noise.
Tables~\ref{tab:explicit-methods} and~\ref{tab:direct-methods} compare explicit-retrieval and direct-reading methods.
\emph{Agent} distinguishes Single and Multi by runtime reasoning roles and collaboration.
\emph{Reasoning} distinguishes fixed workflows from feedback loops: Single-pass covers a single chain of thought, structured decoding, or a fixed multi-agent pipeline; Multi-step revises judgments and actions through repeated feedback from intermediate results, tool observations, or verification.
Multi-step may acquire new evidence through feedback (Iterative acquisition) or only reread and verify existing evidence, as in ViDoRAG's One-shot acquisition \citep{ref18}.
\emph{Learning} records added generator-side training: Pretraining develops document representations and cross-modal alignment; SFT uses answer, evidence, or action demonstrations; RL optimizes policies through answer, evidence, or process rewards.

\begin{figure}[t]
  \centering
  \includegraphics[width=0.9\columnwidth]{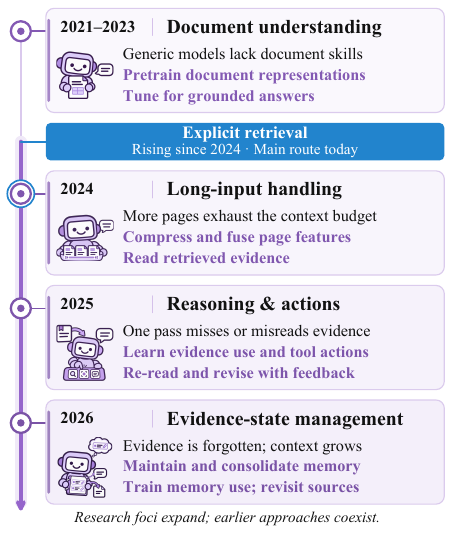}
  \caption{Four stages of generation and learning: bottlenecks in black, technical responses in purple.}
  \label{fig:generation-evolution}
  \vspace{-15pt}
\end{figure}

\begin{table*}[!t]
\centering
\captionsetup{position=top,skip=3pt}
\caption{Representative benchmarks grouped by evaluation goal and input scale. Full details: Appendix~\ref{app:benchmarks}, Table~\ref{tab:benchmark-catalog}.}
\label{tab:benchmarks}
\begingroup
\fontsize{7.8}{8.8}\selectfont
\setlength{\tabcolsep}{1.3pt}
\renewcommand{\arraystretch}{0.90}
\setlength{\extrarowheight}{0pt}
\setlength{\aboverulesep}{1.0pt}
\setlength{\belowrulesep}{1.0pt}
\resizebox{\textwidth}{!}{%
\begin{tabular}{@{}l*{11}{c}@{}}
\toprule
\multicolumn{3}{c}{\textbf{Benchmark}} & \multicolumn{2}{c}{\textbf{Data Size}} & \multicolumn{4}{c}{\textbf{Question Type}} & \multicolumn{3}{c}{\textbf{Evidence and Evaluation}} \\
\cmidrule(lr){1-3}\cmidrule(lr){4-5}\cmidrule(lr){6-9}\cmidrule(l){10-12}
\textbf{Name} & \textbf{Venue} & \textbf{Domain} & \shortstack{\textbf{\#Pages}} & \textbf{\#QA} & \textbf{MD} & \textbf{MP} & \textbf{SP} & \textbf{Unans.} & \textbf{Source} & \textbf{Gran.} & \textbf{Evaluation} \\
\midrule
\multicolumn{12}{@{}l@{}}{\textit{Retrieval-focused benchmarks}} \\
\addlinespace[0.8pt]
\textbf{ViDoRe}~\citeyearpar{ref7}~\MethodDate{24.06} & ICLR'25 & Open & 1.0k & 3.81k & \MethodNo{} & \MethodNo{} & \MethodYes{} & \MethodNo{} & TXT/L/C/TAB/I & P & Ans. Ret. \\
\textbf{REAL-MM-RAG}~\citeyearpar{ref88}~\MethodDate{25.02} & ACL'25 & Open & 2.1k & 4.6k & \MethodNo{} & \MethodNo{} & \MethodYes{} & \MethodNo{} & TXT/C/TAB/I & P & Ans. Ret. \\
\textbf{ViDoRe V2}~\citeyearpar{ref82}~\MethodDate{25.03} & arXiv'25 & Open & 1.1k & 327 & \MethodYes{} & \MethodYes{} & \MethodYes{} & \MethodNo{} & TXT/L/C/TAB/I & P & Ret. \\
\textbf{MMDocIR}~\citeyearpar{ref17}~\MethodDate{25.01} & EMNLP'25 & Closed & 65.1 & 1.66k & \MethodNo{} & \MethodYes{}(18.9\%) & \MethodYes{}(81.1\%) & \MethodNo{} & TXT/L/C/TAB/I & P, E & Ret. Loc. \\
\textbf{CMDR-Bench}~\citeyearpar{ref79}~\MethodDate{26.07} & ECCV'26 & Closed & 183.5 & 800 & \MethodNo{} & \MethodYes{} & -- & \MethodNo{} & TXT/L/C/TAB/I & P & Ret. \\
\midrule
\multicolumn{12}{@{}l@{}}{\textit{Answer-quality benchmarks: single-page inputs}} \\
\addlinespace[0.8pt]
\textbf{DocVQA}~\citeyearpar{mathew2021docvqa}~\MethodDate{20.07} & WACV'21 & Closed & 1.0 & 50k & \MethodNo{} & \MethodNo{} & \MethodYes{} & \MethodNo{} & TXT/L/C/TAB/I & -- & Ans. \\
\textbf{InfographicVQA}~\citeyearpar{ref62}~\MethodDate{21.04} & WACV'22 & Closed & 1.0 & 30k & \MethodNo{} & \MethodNo{} & \MethodYes{} & \MethodNo{} & TXT/L/C/TAB/I & -- & Ans. \\
\textbf{ChartQA}~\citeyearpar{ref75}~\MethodDate{22.03} & ACL'22 & Closed & 1.0 & 32.7k & \MethodNo{} & \MethodNo{} & \MethodYes{} & \MethodNo{} & C & -- & Ans. \\
\textbf{DOGR-Bench}~\citeyearpar{ref15}~\MethodDate{24.11} & ICCV'25 & Closed & 1.0 & 3.6k & \MethodNo{} & \MethodNo{} & \MethodYes{} & \MethodNo{} & TXT/L/C/I & E & Ans. Loc. \\
\midrule
\multicolumn{12}{@{}l@{}}{\textit{Answer-quality benchmarks: multipage inputs or input size unreported}} \\
\addlinespace[0.8pt]
\textbf{TAT-DQA}~\citeyearpar{ref86}~\MethodDate{22.07} & MM'22 & Closed & 1.1 & 16.6k & \MethodNo{} & \MethodYes{} & \MethodYes{} & \MethodNo{} & TXT/L/TAB & -- & Ans. \\
\textbf{SlideVQA}~\citeyearpar{tanaka2023slidevqa}~\MethodDate{23.01} & AAAI'23 & Closed & 20.0 & 14.5k & \MethodNo{} & \MethodYes{}(13.9\%) & \MethodYes{}(86.1\%) & \MethodNo{} & TXT/L/C/TAB/I & P & Ans. Ret. \\
\textbf{MP-DocVQA}~\citeyearpar{tito2023hierarchical}~\MethodDate{22.12} & PR'23 & Closed & 8.3 & 46.2k & \MethodNo{} & \MethodNo{} & \MethodYes{} & \MethodNo{} & TXT/L/C/TAB/I & P & Ans. Ret. \\
\textbf{DUDE}~\citeyearpar{ref3}~\MethodDate{23.05} & ICCV'23 & Closed & 5.72 & 41.5k & \MethodNo{} & \MethodYes{}(2.1\%) & \MethodYes{}(85.2\%) & \MethodYes{}(12.7\%) & TXT/L/C/TAB/I & -- & Ans. \\
\textbf{MMLongBench-Doc}~\citeyearpar{ma2024mmlongbench}~\MethodDate{24.07} & NIPS'24 & Closed & 47.5 & 1.1k & \MethodNo{} & \MethodYes{}(33.7\%) & \MethodYes{}(45.7\%) & \MethodYes{}(20.6\%) & TXT/L/C/TAB/I & P & Ans. Ret. \\
\textbf{Wiki-/Paper-VISA}~\citeyearpar{ref13}~\MethodDate{24.12} & ACL'25 & Open & -- & 90k; 102.2k & \MethodNo{} & \MethodNo{} & \MethodYes{} & \MethodYes{} & TXT/TAB/I & P, E & Ans. Ret. Loc. \\
\textbf{M-LongDoc}~\citeyearpar{ref16}~\MethodDate{24.11} & EMNLP'25 & Closed & 210.8 & 0.9k & \MethodNo{} & \MethodNo{} & \MethodYes{} & \MethodNo{} & TXT/C/TAB/I & P & Ans. Ret. \\
\textbf{DVisRAG}~\citeyearpar{ref89}~\MethodDate{26.02} & CVPR'26 & Open & -- & 367.6k & \MethodNo{} & \MethodYes{} & \MethodYes{} & \MethodNo{} & TXT/L/C/TAB/I & P & Ans. Ret. Rob. \\
\textbf{ViDoRe V3}~\citeyearpar{ref83}~\MethodDate{26.01} & ACL'26 & Open & -- & 3.1k & \MethodYes{} & \MethodYes{} & \MethodYes{} & -- & TXT/L/C/TAB/I & P, E & Ans. Ret. Loc. \\
\textbf{SynthDocBench}~\citeyearpar{ref71}~\MethodDate{26.07} & COLM'26 & Closed & 51.1 & 1.8k & \MethodNo{} & \MethodYes{} & \MethodYes{} & \MethodNo{} & TXT/C/TAB & E & Ans. Ret. \\
\textbf{DocScope}~\citeyearpar{ref113}~\MethodDate{26.05} & NIPS'26 & Closed & 51.3 & 1.12k & \MethodNo{} & \MethodYes{}(57.7\%) & \MethodYes{}(35.3\%) & \MethodYes{}(6.9\%) & TXT/C/TAB/I/L & P, E & Ans. Ret. Loc. \\
\bottomrule
\end{tabular}%
}
\par\vspace{1.5pt}\nointerlineskip
\begin{minipage}{\linewidth}
\fontsize{7.5}{8.5}\selectfont Open/Closed: cross-document corpus/given document. \#Pages: mean input pages (document mean if per-question statistics are unavailable); Open uses the full corpus. \#QA: questions/queries; --: unreported/inapplicable. MD/MP/SP: multi-document/multi-page/single-page evidence; Unans.: unanswerable. TXT/L/C/TAB/I: text/layout/charts/tables/images; P/E: Page/Element.
\end{minipage}
\endgroup
\vspace{-15pt}
\end{table*}

\textbf{Evolution of research focus (Fig.~\ref{fig:generation-evolution}).}
In 2021--2023, pretraining and task fine-tuning established document understanding and grounded answering \citep{tanaka2023slidevqa,tito2023hierarchical,ref67,ref76}.
In 2024, growing page and token counts motivated \emph{long-input handling}: GRAM uses document tokens for cross-page communication \citep{ref117}, while mPLUG-DocOwl2 compresses high-resolution visual features \citep{ref26}.
VisRAG and SV-RAG read retrieved pages \citep{yu2025visrag,chen2024svrag}, while Retrieval-aware Tuning fine-tunes generators with evidence and distractor pages \citep{ref16}.
In 2025, single-pass omissions and misreadings motivated \emph{evidence-guided learning, feedback reasoning, and tool actions}.
ViDoRAG uses cross-role feedback to reread existing candidates \citep{ref18}; DocR1 improves localization and answering with evidence-page rewards \citep{ref27}; and VRAG-RL uses SFT and RL to train VLMs to alternate among search, regional inspection, and answering \citep{wang2026vragrl}.
In 2026, feedback reasoning incorporates \emph{evidence-state management} to reduce forgetting and context accumulation.
DocMemo provides evidence and question memory \citep{ref148}; AWM rewards memory maintenance for answering \citep{ref96}; and SCORE retains provenance, reloads source pages, and reorganizes context before answering \citep{ref150}.

Within feedback loops, Sec.~\ref{sec:retrieval} examines how evidence states guide acquisition; this section covers evidence integration, answer verification, and feedback on remaining gaps.
Organization and learning should be evaluated jointly by answer quality, evidence support, and end-to-end cost.

\subsection{Benchmarks and Evaluation}
\label{sec:evaluation}

\textbf{Evaluation dimensions.}
Table~\ref{tab:benchmarks} compares benchmarks by scale, evidence requirements, and evaluation capabilities.
\emph{Granularity} denotes the verification level supported by public evidence annotations; \emph{Evaluation} identifies assessable capabilities.
Ans. measures answer correctness; Ret. evaluates evidence recall and ranking; Loc. measures spatial alignment between predicted and annotated regions.
Eff. covers parsing, indexing, and online inference costs; Rob. assesses performance under visual degradation or interference.
Joint reporting under comparable inputs and budgets helps relate capability gains to end-to-end costs.

\begin{figure}[t]
  \centering
  \includegraphics[width=0.9\columnwidth]{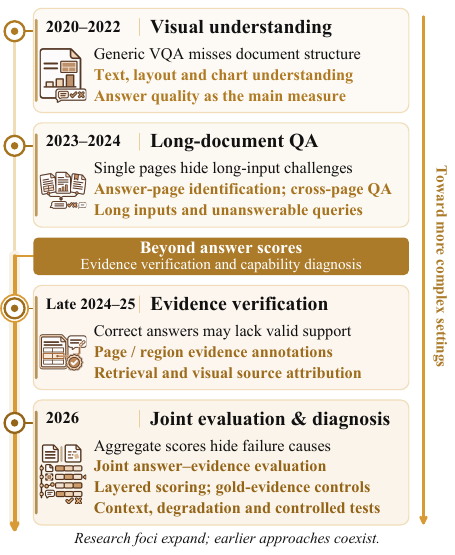}
  \caption{Four stages of benchmarks and evaluation: bottlenecks in black, technical responses in amber.}
  \label{fig:evaluation-evolution}
  \vspace{-15pt}
\end{figure}

\textbf{Evolution of research focus (Fig.~\ref{fig:evaluation-evolution}).}
In 2020--2022, DocVQA, InfographicVQA, and ChartQA benchmarked text, layout, and chart understanding \citep{mathew2021docvqa,ref62,ref75}.
In 2023--2024, longer inputs shifted evaluation toward page identification and cross-page integration \citep{tanaka2023slidevqa,tito2023hierarchical}.
DUDE adds complex and unanswerable questions \citep{ref3}, while MMLongBench-Doc and M-LongDoc assess long-document multimodal understanding \citep{ma2024mmlongbench,ref16}.
From late 2024 through 2025, RAG prompted evidence and provenance checks beyond answer scores: VISA grounds answers in source-page regions \citep{ref13}, while MMDocIR evaluates page- and element-level retrieval \citep{ref17}.
Work in 2026 targets \emph{joint evaluation and capability diagnosis in complex settings}.
ViDoRe V3 links retrieval and QA on a shared corpus, with annotated relevant pages as a control \citep{ref83}; DocScope diagnoses bottlenecks through page, region, fact, and answer scores under varying evidence conditions \citep{ref113}.
CMDR-Bench requires cross-page context for retrieval \citep{ref79}; DVisRAG evaluates retrieval, generation, and end-to-end robustness under visual degradation \citep{ref89}; and SynthDocBench uses controlled synthesis and structured evidence metadata to diagnose capabilities across document conditions and question types \citep{ref71}.
Across stages, tasks expand to longer documents, cross-page and multi-element evidence, multihop reasoning, and diverse applications.

\begin{figure*}[t]
  \centering
  \includegraphics[width=0.95\textwidth]{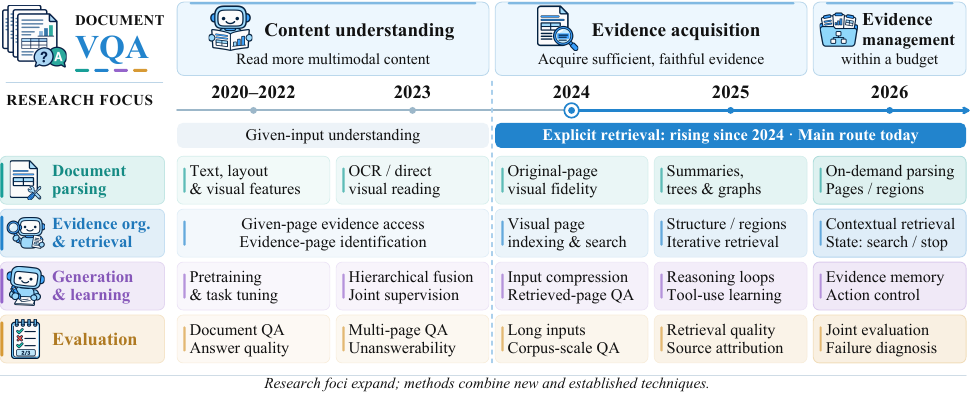}
  \caption{Research foci and technical evolution across four components of document VQA. Five periods span three overarching foci: content understanding, evidence acquisition, and sustained evidence management.}
  \label{fig:cross-component-evolution}
  \vspace{-10pt}
\end{figure*}

\section{Research Trends and Future Directions}
\label{sec:research-trends}
Building on component analyses, we examine how their interactions shape priorities and outline challenges and future directions at two levels: improving individual-question answering and extending the task toward continual document understanding.

\subsection{Cross Component Evolution}
\label{sec:cross-component}

Document VQA research has broadened from content understanding to evidence acquisition and sustained evidence management (Fig.~\ref{fig:cross-component-evolution}).
In 2020--2023, multipage representations and cross-page fusion extended single-page understanding \citep{tanaka2023slidevqa,tito2023hierarchical}.
In 2024--2025, costly exhaustive reading motivated sufficient, high-fidelity evidence acquisition.
Visual page indexing and reading after retrieval in 2024 helped establish explicit retrieval as the main route \citep{ref7,yu2025visrag}.
Structural relations, fine-grained selection, and feedback-driven acquisition improved coverage and use of local details in 2025 \citep{wang2026vragrl,ref36}.
In 2026, attention extends to maintaining evidence states, identifying gaps, and coordinating evidence use and verification across rounds under budget constraints \citep{ref96,ref148,ref150}.

Bottlenecks also propagate across components.
Retrieval narrows inputs but risks omissions, motivating structure-aware acquisition and on-demand parsing to preserve evidence relations and details \citep{ref36,ref104}.
Feedback makes parsing and retrieval schedulable actions, with generation maintaining evidence states and guiding rereading, acquisition, and stopping \citep{wang2026vragrl,ref148}.
Evaluation must diagnose failures through answers and evidence \citep{ref113} and assess gains against end-to-end cost.
Methods may update selected components while retaining established techniques elsewhere; evolution thus reflects broadening priorities rather than wholesale replacement.

\subsection{Challenges and Future Directions}
\label{sec:future-directions}

For individual questions, we identify three research directions.
\textbf{Evidence sufficiency.} Cross-page context and region extraction improve acquisition \citep{ref79,ref104}, but local relevance and precise localization do not ensure sufficient support.
Future work should construct question-driven evidence sets and assess their sufficiency.
\textbf{Verifiable and correctable document understanding.} Cross-round memory and multilevel evaluation support process verification \citep{ref113,ref148}, but correct answers may mask unreliable intermediate judgments, and early errors can propagate.
Future work should link source tracing, process verification, and error correction.
\textbf{Budget-aware coordination.} Action learning and dynamic retrieval improve adaptive acquisition \citep{wang2026vragrl}, but component gains may not improve the full pipeline.
Future work should coordinate decisions across components using evidence states and budget constraints to balance answer quality and end-to-end cost.

Document use often involves follow-up questions and reading across tasks, motivating \textbf{memory reuse across questions} and \textbf{continual, experience-driven strategy improvement}.
The former reuses evidence and interaction states for subsequent questions; the latter turns cross-task experience into transferable retrieval, reading, and verification strategies.
Appendix~\ref{app:research-opportunities} details challenges, research opportunities, and task extensions.

\section{Conclusion}
\label{sec:conclusion}

This survey decomposes long-document VQA into four components and links a fine-grained taxonomy with task bottlenecks, research motivations, and technical evolution.
Emphasizing explicit retrieval while covering direct reading, we trace expanding priorities from multimodal understanding to sufficient, high-fidelity evidence acquisition and sustained management under limited budgets.
Future directions include evidence sufficiency assessment, verifiable and correctable understanding, and budget-aware coordination for individual questions, alongside memory reuse across questions and strategy improvement across tasks toward continual document understanding.

\section*{Limitations}

This survey focuses on document VQA as a higher-level understanding task, particularly evidence acquisition and use in long documents.
Foundational tasks such as text and layout recognition and field and entity extraction are covered only where they affect question answering, without being reviewed in detail.

\section*{Use of AI-Assisted Tools}

We used ChatGPT for language editing and LaTeX formatting. We reviewed and edited the AI-assisted content and take full responsibility for the final manuscript.

\section*{Conflicts of Interest}

We declare no conflicts of interest.

\bibliography{references}

@article{tanaka2023slidevqa,
  title={Slidevqa: A dataset for document visual question answering on multiple images},
  author={Tanaka, Ryota and Nishida, Kyosuke and Nishida, Kosuke and Hasegawa, Taku and Saito, Itsumi and Saito, Kuniko},
  journal={Proceedings of the AAAI Conference on Artificial Intelligence},
  volume={37},
  number={11},
  pages={13636--13645},
  year={2023}
}

@article{tito2023hierarchical,
  title={Hierarchical multimodal transformers for multipage docvqa},
  author={Tito, Rub{\`e}n and Karatzas, Dimosthenis and Valveny, Ernest},
  journal={Pattern Recognition},
  volume={144},
  pages={109834},
  year={2023}
}

@article{ma2024mmlongbench,
  title={Mmlongbench-doc: Benchmarking long-context document understanding with visualizations},
  author={Ma, Yubo and Zang, Yuhang and Chen, Liangyu and Chen, Meiqi and Jiao, Yizhu and Li, Xinze and Lu, Xinyuan and Liu, Ziyu and Ma, Yan and Dong, Xiaoyi and Zhang, Pan and Pan, Liangming and Jiang, Yu-Gang and Wang, Jiaqi and Cao, Yixin and Sun, Aixin},
  journal={Advances in Neural Information Processing Systems},
  volume={37},
  pages={95963--96010},
  year={2024}
}

@inproceedings{ma2024unifying,
  title={Unifying multimodal retrieval via document screenshot embedding},
  author={Ma, Xueguang and Lin, Sheng-Chieh and Li, Minghan and Chen, Wenhu and Lin, Jimmy},
  booktitle={Proceedings of the 2024 Conference on Empirical Methods in Natural Language Processing},
  pages={6492--6505},
  year={2024}
}

@inproceedings{yu2025visrag,
  title={Visrag: Vision-based retrieval-augmented generation on multi-modality documents},
  author={Yu, Shi and Tang, Chaoyue and Xu, Bokai and Cui, Junbo and Ran, Junhao and Yan, Yukun and Liu, Zhenghao and Wang, Shuo and Han, Xu and Liu, Zhiyuan and Sun, Maosong},
  booktitle={International Conference on Learning Representations},
  volume={2025},
  pages={21074--21098},
  year={2025}
}

@article{chen2024svrag,
  title={SV-RAG: LoRA-contextualizing adaptation of MLLMs for long document understanding},
  author={Chen, Jian and Zhang, Ruiyi and Zhou, Yufan and Yu, Tong and Dernoncourt, Franck and Gu, Jiuxiang and Rossi, Ryan A. and Chen, Changyou and Sun, Tong},
  journal={arXiv preprint arXiv:2411.01106},
  year={2024}
}

@inproceedings{cho2025m3docvqa,
  title={M3docvqa: Multi-modal multi-page multi-document understanding},
  author={Cho, Jaemin and Mahata, Debanjan and Irsoy, Ozan and He, Yujie and Bansal, Mohit},
  booktitle={2025 IEEE/CVF International Conference on Computer Vision Workshops (ICCVW)},
  pages={6237--6247},
  year={2025},
  organization={IEEE}
}

@article{wang2026vragrl,
  title={Vrag-rl: Empower vision-perception-based rag for visually rich information understanding via iterative reasoning with reinforcement learning},
  author={Wang, Qiuchen and Ding, Ruixue and Zeng, Yu and Chen, Zehui and Chen, Lin and Wang, Shihang and Xie, Pengjun and Huang, Fei and Zhao, Feng},
  journal={Advances in Neural Information Processing Systems},
  volume={38},
  pages={57133--57160},
  year={2026}
}

@inproceedings{mathew2021docvqa,
  title={Docvqa: A dataset for vqa on document images},
  author={Mathew, Minesh and Karatzas, Dimosthenis and Jawahar, C. V.},
  booktitle={2021 IEEE Winter Conference on Applications of Computer Vision (WACV)},
  pages={2199--2208},
  year={2021},
  organization={IEEE}
}

@inproceedings{huang2022layoutlmv3,
  title={Layoutlmv3: Pre-training for document ai with unified text and image masking},
  author={Huang, Yupan and Lv, Tengchao and Cui, Lei and Lu, Yutong and Wei, Furu},
  booktitle={Proceedings of the 30th ACM international conference on multimedia},
  pages={4083--4091},
  year={2022}
}

@inproceedings{gao2026scaling,
  title={Scaling beyond context: A survey of multimodal retrieval-augmented generation for document understanding},
  author={Gao, Sensen and Zhao, Shanshan and Jiang, Xu and Duan, Lunhao and Chng, Yong Xien and Chen, Qing-Guo and Luo, Weihua and Zhang, Kaifu and Bian, Jia-Wang and Gong, Mingming},
  booktitle={Proceedings of the 64th Annual Meeting of the Association for Computational Linguistics (Volume 1: Long Papers)},
  pages={4458--4489},
  year={2026}
}

@article{yan2026unlocking,
  title={Unlocking Multimodal Document Intelligence: From Current Triumphs to Future Frontiers of Visual Document Retrieval},
  author={Yan, Yibo and Huo, Jiahao and Feng, Guanbo and Ou, Mingdong and Cao, Yi and Zou, Xin and Liu, Shuliang and Lyu, Yuanhuiyi and Huang, Yu and Li, Jungang and Zheng, Kening and Zheng, Xu and Yu, Philip S. and Kwok, James and Hu, Xuming},
  journal={arXiv preprint arXiv:2602.19961},
  year={2026}
}

@article{ding2026survey,
  title={A survey on MLLM-based visually rich document understanding: Methods, challenges, and emerging trends},
  author={Ding, Yihao and Luo, Siwen and Dai, Yue and Jiang, Yanbei and Li, Zechuan and Sun, Qiang and Martin, Geoffrey and Liu, Wei and Peng, Yifan},
  journal={Findings of the Association for Computational Linguistics: ACL 2026},
  pages={13319--13340},
  year={2026}
}

@article{barboule2025survey,
  title={Survey on question answering over visually rich documents: Methods, challenges, and trends},
  author={Barboule, Camille and Piwowarski, Benjamin and Chabot, Yoan},
  journal={arXiv preprint arXiv:2501.02235},
  year={2025}
}

@inproceedings{dong2026docresearcher,
  title={Doc-researcher: A unified system for multimodal document parsing and deep research},
  author={Dong, Kuicai and Huang, Shurui and Ye, Fangda and Han, Wei and Zhang, Zhi and Li, Dexun and Li, Wenjun and Yang, Qu and Wang, Gang and Wang, Yichao and Zhang, Chen and Liu, Yong},
  booktitle={Proceedings of the ACM Web Conference 2026},
  pages={2349--2360},
  year={2026}
}

@inproceedings{ref3,
  title = {Document understanding dataset and evaluation (dude)},
  author = {Van Landeghem, Jordy and Tito, Rubèn and Borchmann, Łukasz and Pietruszka, Michał and Joziak, Pawel and Powalski, Rafal and Jurkiewicz, Dawid and Coustaty, Mickael and Anckaert, Bertrand and Valveny, Ernest and Blaschko, Matthew and Moens, Sien and Stanislawek, Tomasz},
  booktitle = {Proceedings of the IEEE/CVF international conference on computer vision},
  year = {2023},
  pages = {19528--19540}
}

@inproceedings{ref7,
  title = {Colpali: Efficient document retrieval with vision language models},
  author = {Faysse, Manuel and Sibille, Hugues and Wu, Tony and Omrani, Bilel and Viaud, Gautier and Hudelot, Céline and Colombo, Pierre},
  booktitle = {International Conference on Learning Representations},
  year = {2025},
  volume = {2025},
  pages = {61424--61449}
}

@inproceedings{ref10,
  title = {Visdom: Multi-document qa with visually rich elements using multimodal retrieval-augmented generation},
  author = {Suri, Manan and Mathur, Puneet and Dernoncourt, Franck and Goswami, Kanika and Rossi, Ryan A. and Manocha, Dinesh},
  booktitle = {Proceedings of the 2025 Conference of the Nations of the Americas Chapter of the Association for Computational Linguistics: Human Language Technologies (Volume 1: Long Papers)},
  year = {2025},
  pages = {6088--6109}
}

@inproceedings{ref11,
  title = {Vdocrag: Retrieval-augmented generation over visually-rich documents},
  author = {Tanaka, Ryota and Iki, Taichi and Hasegawa, Taku and Nishida, Kyosuke and Saito, Kuniko and Suzuki, Jun},
  booktitle = {2025 IEEE/CVF Conference on Computer Vision and Pattern Recognition (CVPR)},
  year = {2025},
  pages = {24827--24837},
  organization = {IEEE}
}

@inproceedings{ref13,
  title = {Visa: Retrieval augmented generation with visual source attribution},
  author = {Ma, Xueguang and Zhuang, Shengyao and Koopman, Bevan and Zuccon, Guido and Chen, Wenhu and Lin, Jimmy},
  booktitle = {Proceedings of the 63rd Annual Meeting of the Association for Computational Linguistics (Volume 1: Long Papers)},
  year = {2025},
  pages = {30154--30169}
}

@inproceedings{ref15,
  title = {Dogr: Towards versatile visual document grounding and referring},
  author = {Zhou, Yinan and Chen, Yuxin and Lin, Haokun and Wu, Yichen and Yang, Shuyu and Qi, Zhongang and Ma, Chen and Zhu, Li},
  booktitle = {2025 IEEE/CVF International Conference on Computer Vision (ICCV)},
  year = {2025},
  pages = {3596--3606},
  organization = {IEEE}
}

@inproceedings{ref16,
  title = {M-longdoc: A benchmark for multimodal super-long document understanding and a retrieval-aware tuning framework},
  author = {Chia, Yew Ken and Cheng, Liying and Chan, Hou Pong and Song, Maojia and Liu, Chaoqun and Aljunied, Mahani and Poria, Soujanya and Bing, Lidong},
  booktitle = {Proceedings of the 2025 Conference on Empirical Methods in Natural Language Processing},
  year = {2025},
  pages = {9244--9261}
}

@inproceedings{ref17,
  title = {MMDocIR: Benchmarking multimodal retrieval for long documents},
  author = {Dong, Kuicai and Chang, Yujing and Deik, Derrick Goh Xin and Li, Dexun and Tang, Ruiming and Liu, Yong},
  booktitle = {Proceedings of the 2025 Conference on Empirical Methods in Natural Language Processing},
  year = {2025},
  pages = {30959--30993}
}

@inproceedings{ref18,
  title = {Vidorag: Visual document retrieval-augmented generation via dynamic iterative reasoning agents},
  author = {Wang, Qiuchen and Ding, Ruixue and Chen, Zehui and Wu, Weiqi and Wang, Shihang and Xie, Pengjun and Zhao, Feng},
  booktitle = {Proceedings of the 2025 Conference on Empirical Methods in Natural Language Processing},
  year = {2025},
  pages = {9124--9145}
}

@inproceedings{ref25,
  title = {Docvlm: Make your vlm an efficient reader},
  author = {Nacson, Mor Shpigel and Aberdam, Aviad and Ganz, Roy and Ben Avraham, Elad and Golts, Alona and Kittenplon, Yair and Mazor, Shai and Litman, Ron},
  booktitle = {Proceedings of the IEEE/CVF Conference on Computer Vision and Pattern Recognition},
  year = {2025},
  pages = {29005--29015}
}

@inproceedings{ref26,
  title = {mplug-docowl2: High-resolution compressing for ocr-free multi-page document understanding},
  author = {Hu, Anwen and Xu, Haiyang and Zhang, Liang and Ye, Jiabo and Yan, Ming and Zhang, Ji and Jin, Qin and Huang, Fei and Zhou, Jingren},
  booktitle = {Proceedings of the 63rd Annual Meeting of the Association for Computational Linguistics (Volume 1: Long Papers)},
  year = {2025},
  pages = {5817--5834}
}

@article{ref27,
  title = {Docr1: Evidence page-guided grpo for multi-page document understanding},
  author = {Xiong, Junyu and Wang, Yonghui and Zhao, Weichao and Liu, Chenyu and Yin, Bing and Zhou, Wengang and Li, Houqiang},
  journal = {Proceedings of the AAAI Conference on Artificial Intelligence},
  year = {2026},
  volume = {40},
  number = {13},
  pages = {11178--11186}
}

@article{ref29,
  title = {DocSeeker: Structured visual reasoning with evidence grounding for long document understanding},
  author = {Yan, Hao and Liu, Yuliang and Liu, Xingchen and Zhang, Yuyi and Liao, Minghui and Wu, Jihao and Chen, Wei and Bai, Xiang},
  journal = {arXiv preprint arXiv:2604.12812},
  year = {2026}
}

@inproceedings{ref30,
  title = {Doc-react: Multi-page heterogeneous document question-answering},
  author = {Wu, Junda and Xia, Yu and Yu, Tong and Chen, Xiang and Harsha, Sai Sree and Maharaj, Akash V. and Zhang, Ruiyi and Bursztyn, Victor and Kim, Sungchul and Rossi, Ryan A. and McAuley, Julian and Li, Yunyao and Sinha, Ritwik},
  booktitle = {Proceedings of the 63rd Annual Meeting of the Association for Computational Linguistics (Volume 2: Short Papers)},
  year = {2025},
  pages = {67--78}
}

@article{ref32,
  title = {Mdocagent: A multi-modal multi-agent framework for document understanding},
  author = {Han, Siwei and Xia, Peng and Zhang, Ruiyi and Sun, Tong and Li, Yun and Zhu, Hongtu and Yao, Huaxiu},
  journal = {arXiv preprint arXiv:2503.13964},
  year = {2025}
}

@inproceedings{ref34,
  title = {SimpleDoc: Multi‐Modal Document Understanding with Dual‐Cue Page Retrieval and Iterative Refinement},
  author = {Jain, Chelsi and Wu, Yiran and Zeng, Yifan and Liu, Jiale and Dai, Shengyu and Shao, Zhenwen and Wu, Qingyun and Wang, Huazheng},
  booktitle = {Proceedings of the 2025 Conference on Empirical Methods in Natural Language Processing},
  year = {2025},
  pages = {28398--28415}
}

@article{ref42,
  title = {Rag-anything: All-in-one rag framework},
  author = {Guo, Zirui and Ren, Xubin and Xu, Lingrui and Zhang, Jiahao and Huang, Chao},
  journal = {arXiv preprint arXiv:2510.12323},
  year = {2025}
}

@article{ref44,
  title = {DocPrune: Efficient Document Question Answering via Background, Question, and Comprehension-aware Token Pruning},
  author = {Choi, Joonmyung and Lee, Sanghyeok and Kim, Jongha and Kim, Sehyung and Ko, Dohwan and Kil, Jihyung and Kim, Hyunwoo J.},
  journal = {arXiv preprint arXiv:2604.22281},
  year = {2026}
}

@article{ref46,
  title = {CogDoc: Towards unified thinking in documents},
  author = {Xu, Qixin and Wang, Haozhe and Liu, Che and Lin, Fangzhen and Chen, Wenhu},
  journal = {arXiv preprint arXiv:2512.12658},
  year = {2025}
}

@inproceedings{ref48,
  title = {Doclens: A tool-augmented multi-agent framework for long visual document understanding},
  author = {Zhu, Dawei and Meng, Rui and Chen, Jiefeng and Li, Sujian and Pfister, Tomas and Yoon, Jinsung},
  booktitle = {Proceedings of the 64th Annual Meeting of the Association for Computational Linguistics (Volume 1: Long Papers)},
  year = {2026},
  pages = {26804--26829}
}

@inproceedings{ref49,
  title = {MAB-DQA: Addressing query aspect importance in document question answering with multi-armed bandits},
  author = {Xiang, Yixin and Ma, Yunshan and Du, Xiaoyu and Chen, Yibing and Zhang, Yanxin and Tang, Jinhui},
  booktitle = {Proceedings of the 64th Annual Meeting of the Association for Computational Linguistics (Volume 1: Long Papers)},
  year = {2026},
  pages = {22973--22992}
}

@inproceedings{ref50,
  title = {Doc-V*: Coarse-to-Fine Interactive Visual Reasoning for Multi-Page Document VQA},
  author = {Zheng, Yuanlei and Fu, Pei and Li, Hang and Wang, Ziyang and Zhang, Yuyi and Ruan, Wenyu and Zhang, Xiaojin and Wei, Zhongyu and Luo, Zhenbo and Luan, Jian and Chen, Wei and Bai, Xiang},
  booktitle = {Proceedings of the 64th Annual Meeting of the Association for Computational Linguistics (Volume 1: Long Papers)},
  year = {2026},
  pages = {45901--45923}
}

@inproceedings{ref51,
  title = {MM-Doc-R1: Training Agents for Long Document Visual Question Answering through Multi-turn Reinforcement Learning},
  author = {Lin, Jiahang and Hu, Kai and Wang, Binghai and Zhou, Yuhao and Xi, Zhiheng and Guo, Honglin and Liu, Shichun and Wang, Junzhe and Dou, Shihan and Zhou, Enyu and Yan, Hang and Han, Zhenhua and Gui, Tao and Zhang, Qi and Huang, Xuanjing},
  booktitle = {Findings of the Association for Computational Linguistics: ACL 2026},
  year = {2026},
  pages = {29770--29783}
}

@article{ref52,
  title = {Docdancer: Towards agentic document-grounded information seeking},
  author = {Zhang, Qintong and Lv, Xinjie and Wu, Jialong and Li, Baixuan and Tao, Zhengwei and Yan, Guochen and Zhang, Huanyao and Wang, Bin and Xu, Jiahao and Mi, Haitao and Zhang, Wentao},
  journal = {arXiv preprint arXiv:2601.05163},
  year = {2026}
}

@article{ref53,
  title = {VimRAG: Navigating Massive Visual Context in Retrieval-Augmented Generation via Multimodal Memory Graph},
  author = {Wang, Qiuchen and Wang, Shihang and Zeng, Yu and Zhang, Qiang and Zhang, Fanrui and Guo, Zhuoning and Zhang, Bosi and Huang, Wenxuan and Chen, Lin and Chen, Zehui and Xie, Pengjun and Ding, Ruixue},
  journal = {arXiv preprint arXiv:2602.12735},
  year = {2026}
}

@article{ref56,
  title = {MAGE-RAG: Multigranular Adaptive Graph Evidence for Agentic Multimodal RAG in Long-Document QA},
  author = {Zuo, Yilong and Li, Xunkai and Yuan, Jing and Dai, Qiangqiang and Qin, Hongchao and Li, Ronghua},
  journal = {arXiv preprint arXiv:2606.15906},
  year = {2026}
}

@article{ref59,
  title = {Doc-REFRAG: Rethinking Multimodal Document Retrieval-Augmented Generation},
  author = {Hu, Ruofan and Xu, Shengyang and Hong, Minjie and Yang, Xiaoda and Zhou, Sashuai and Lei, Ke and Jin, Tao and Zhao, Zhou},
  journal = {arXiv preprint arXiv:2608.30163},
  year = {2026}
}

@article{ref60,
  title = {DocTrace: Towards Traceable Long Document VQA via Hierarchical Evidence Graph Reasoning},
  author = {Xiang, Le and Guan, Zhicheng and Chen, Hong and Lin, Xiaocong and Lei, Zhenghua and Hu, Teng and He, Bolei and Zeng, Long},
  journal = {arXiv preprint arXiv:2608.03292},
  year = {2026}
}

@inproceedings{ref62,
  title = {Infographicvqa},
  author = {Mathew, Minesh and Bagal, Viraj and Tito, Rubèn and Karatzas, Dimosthenis and Valveny, Ernest and Jawahar, C. V.},
  booktitle = {2022 IEEE/CVF Winter Conference on Applications of Computer Vision (WACV)},
  year = {2022},
  pages = {2582--2591},
  publisher = {IEEE}
}

@inproceedings{ref67,
  title = {Ocr-free document understanding transformer},
  author = {Kim, Geewook and Hong, Teakgyu and Yim, Moonbin and Nam, JeongYeon and Park, Jinyoung and Yim, Jinyeong and Hwang, Wonseok and Yun, Sangdoo and Han, Dongyoon and Park, Seunghyun},
  booktitle = {European Conference on Computer Vision},
  year = {2022},
  pages = {498--517},
  publisher = {Springer Nature Switzerland},
  address = {Cham}
}

@inproceedings{ref70,
  title = {SCoPE VLM: Selective Context Processing for Efficient Document Navigation in Vision-Language Models},
  author = {Lim, Gyubeum and Koo, Yemo and Madisetti, Vijay Krishna},
  booktitle = {Proceedings of the 19th Conference of the European Chapter of the Association for Computational Linguistics (Volume 1: Long Papers)},
  year = {2026},
  pages = {95--140}
}

@inproceedings{ref83,
  title = {Vidore v3: A comprehensive evaluation of retrieval augmented generation in complex real-world scenarios},
  author = {Loison, António and Macé, Quentin and Edy, Antoine and Xing, Victor and Balough, Tom and Moreira, Gabriel de Souza P. and Liu, Bo and Faysse, Manuel and Hudelot, Céline and Viaud, Gautier},
  booktitle = {Proceedings of the 64th Annual Meeting of the Association for Computational Linguistics (Volume 1: Long Papers)},
  year = {2026},
  pages = {16570--16600}
}

@inproceedings{ref86,
  title = {Towards complex document understanding by discrete reasoning},
  author = {Zhu, Fengbin and Lei, Wenqiang and Feng, Fuli and Wang, Chao and Zhang, Haozhou and Chua, Tat-Seng},
  booktitle = {Proceedings of the 30th ACM International Conference on Multimedia},
  year = {2022},
  pages = {4857--4866}
}

@inproceedings{ref87,
  title = {Unifying vision, text, and layout for universal document processing},
  author = {Tang, Zineng and Yang, Ziyi and Wang, Guoxin and Fang, Yuwei and Liu, Yang and Zhu, Chenguang and Zeng, Michael and Zhang, Cha and Bansal, Mohit},
  booktitle = {2023 IEEE/CVF Conference on Computer Vision and Pattern Recognition (CVPR)},
  year = {2023},
  pages = {19254--19264},
  publisher = {IEEE}
}

@inproceedings{ref88,
  title = {Real-mm-rag: A real-world multi-modal retrieval benchmark},
  author = {Wasserman, Navve and Pony, Roi and Naparstek, Oshri and Goldfarb, Adi Raz and Schwartz, Eli and Barzelay, Udi and Karlinsky, Leonid},
  booktitle = {Proceedings of the 63rd Annual Meeting of the Association for Computational Linguistics (Volume 1: Long Papers)},
  year = {2025},
  pages = {31660--31683}
}

@article{ref101,
  title = {RegionRAG: Region-level Retrieval-Augmented Generation for Visual Document Understanding},
  author = {Li, Yinglu and Lu, Zhiying and Liu, Zhihang and Sun, Yiwei and Liu, Chuanbin and Xie, Hongtao},
  journal = {Proceedings of the AAAI Conference on Artificial Intelligence},
  year = {2026},
  volume = {40},
  number = {8},
  pages = {6662--6670}
}

@inproceedings{ref89,
  title = {RobustVisRAG: Causality-Aware Vision-Based Retrieval-Augmented Generation under Visual Degradations},
  author = {Chen, I. and Liu, Yu-Wei and Wu, Tse-Yu and Chiang, Yu-Chien and Yang, Jen-Chieh and Chen, Wei-Ting},
  booktitle = {Proceedings of the IEEE/CVF Conference on Computer Vision and Pattern Recognition},
  year = {2026},
  pages = {40802--40811}
}

@inproceedings{ref92,
  title = {Ureader: Universal ocr-free visually-situated language understanding with multimodal large language model},
  author = {Ye, Jiabo and Hu, Anwen and Xu, Haiyang and Ye, Qinghao and Yan, Ming and Xu, Guohai and Li, Chenliang and Tian, Junfeng and Qian, Qi and Zhang, Ji and Jin, Qin and He, Liang and Lin, Xin and Huang, Fei},
  booktitle = {Findings of the Association for Computational Linguistics: EMNLP 2023},
  year = {2023},
  pages = {2841--2858}
}

@article{ref94,
  title={MoDora: Tree-Based Semi-Structured Document Analysis System},
  author={Xu, Bangrui and Yao, Qihang and Tang, Zirui and Zhou, Xuanhe and He, Yeye and Yu, Shihan and Xu, Qianqian and Wang, Bin and Li, Guoliang and He, Conghui and others},
  journal={Proceedings of the ACM on Management of Data},
  volume={4},
  number={3 (SIGMOD)},
  pages={1--26},
  year={2026},
  publisher={ACM New York, NY, USA}
}

@article{ref96,
  title = {AWM: Answerable Working Memory for Long-Document VQA Agents},
  author = {Zhou, Dongzhuoran and Zhu, Yuqicheng and Liu, Yule and Yang, Zhen and Lu, Rui and Dong, Yuxiao and Tang, Jie and Kharlamov, Evgeny},
  journal = {arXiv preprint arXiv:2608.25618},
  year = {2026}
}

@article{ref104,
  title = {Agenticocr: Parsing only what you need for efficient retrieval-augmented generation},
  author = {Wang, Zhengren and Ma, Dongsheng and Zhong, Huaping and Li, Jiayu and Zhang, Wentao and Wang, Bin and He, Conghui},
  journal = {arXiv preprint arXiv:2602.24134},
  year = {2026}
}

@article{ref113,
  title = {DocScope: Benchmarking Verifiable Reasoning for Trustworthy Long-Document Understanding},
  author = {Feng, Xiang and Zhou, Jiawei and Huang, Zhangfeng and Wang, Kewei and Ye, Shanshan and Hu, Jinxin and Chen, Zulong and Luo, Yong and Zhang, Jing},
  journal = {arXiv preprint arXiv:2605.08888},
  year = {2026}
}

@article{ref115,
  title = {EviProp: Seeded Relevance Diffusion on Chunk-Page Graphs for Long Multimodal Document Retrieval},
  author = {Zhang, Hongwei and Wang, Xiaoman and Ling, Zehui and Zhu, Ruicheng and Zhang, Yue and Cai, Pinlong and Shen, Fuke and Shi, Botian and Wei, Tongquan and Yan, Guohang},
  journal = {arXiv preprint arXiv:2606.08979},
  year = {2026}
}

@inproceedings{ref117,
  title = {Gram: Global reasoning for multi-page vqa},
  author = {Blau, Tsachi and Fogel, Sharon and Ronen, Roi and Golts, Alona and Ganz, Roy and Ben Avraham, Elad and Aberdam, Aviad and Tsiper, Shahar and Litman, Ron},
  booktitle = {Proceedings of the IEEE/CVF conference on computer vision and pattern recognition},
  year = {2024},
  pages = {15598--15607}
}

@inproceedings{ref118,
  title = {Cream: coarse-to-fine retrieval and multi-modal efficient tuning for document vqa},
  author = {Zhang, Jinxu and Yu, Yongqi and Zhang, Yu},
  booktitle = {Proceedings of the 32nd ACM International Conference on Multimedia},
  year = {2024},
  pages = {925--934}
}

@article{ref126,
  title = {Can Retrieval Heads See Images? Multimodal Retrieval Heads in Long-Context Vision-Language Models},
  author = {Li, Aaron Branson Cigres and Wang, Zhaowei and Zhao, Yu and Du, Yiming and Li, Haobo and Ren, Xiyu and Wong, Ginny and See, Simon and Luo, Lishu and Duan, Haodong and Minervini, Pasquale and Song, Yangqiu},
  journal = {arXiv preprint arXiv:2605.27243},
  year = {2026}
}

@inproceedings{ref127,
  title = {LMS-Retrieval: Layout-Aware, Modality-Aware, Structure-Aware Document Retrieval},
  author = {Qin, Man and French, Tim and Liu, Wei},
  booktitle = {International Conference on Document Analysis and Recognition},
  year = {2026},
  pages = {123--139},
  publisher = {Springer Nature Switzerland},
  address = {Cham}
}

@inproceedings{ref35,
  title={Molorag: Bootstrapping document understanding via multi-modal logic-aware retrieval},
  author={Wu, Xixi and Yanchao Tan and Nan Hou and Ruiyang Zhang and Hong Cheng},
  booktitle={Proceedings of the 2025 Conference on Empirical Methods in Natural Language Processing},
  year={2025},
  pages={14035--14056},
  month={November}
}

@inproceedings{ref36,
  title={Lilac: Late interacting in layered component graph for open-domain multimodal multihop retrieval},
  author={Yun, Joohyung and Doyup Lee and Wook-Shin Han},
  booktitle={Proceedings of the 2025 Conference on Empirical Methods in Natural Language Processing},
  year={2025},
  pages={20551--20570},
  month={November}
}

@inproceedings{ref37,
  title={Docagent: An agentic framework for multi-modal long-context document understanding},
  author={Sun, Li and Liu He and Shuyue Jia and Yangfan He and Chenyu You},
  booktitle={Proceedings of the 2025 Conference on Empirical Methods in Natural Language Processing},
  year={2025},
  pages={17712--17727},
  month={November}
}

@article{ref38,
  title={URaG: Unified retrieval and generation in multimodal LLMs for efficient long document understanding},
  author={Shi, Yongxin and Jiapeng Wang and Zeyu Shan and Dezhi Peng and Zening Lin and Lianwen Jin},
  journal={Proceedings of the AAAI Conference on Artificial Intelligence},
  year={2026},
  volume={40},
  number={30},
  pages={25357--25365},
  month={March}
}

@inproceedings{ref40,
  title={Dmap: Human-aligned structural document map for multimodal document understanding},
  author={Fu, Shunliang and Yanxin Zhang and Yixin Xiang and Xiaoyu Du and Jinhui Tang},
  booktitle={Proceedings of the ACM Web Conference 2026},
  year={2026},
  pages={2037--2048},
  month={April}
}

@article{ref41,
  title={Visrag 2.0: Evidence-guided multi-image reasoning in visual retrieval-augmented generation},
  author={Sun, Yubo and Chunyi Peng and Yukun Yan and Shi Yu and Zhenghao Liu and Chi Chen and Zhiyuan Liu and Maosong Sun},
  journal={arXiv preprint arXiv:2510.09733},
  year={2025}
}

@article{ref71,
  title={SynthDocBench: Controlled Benchmark for Long-Context Visual Document Understanding},
  author={Verma, Abhigya and Khyati Mahajan and Amit Kumar Saha and Shruthan Radhakrishna and Sagar Davasam and Vikas Yadav and Sai Rajeswar Mudumba},
  journal={arXiv preprint arXiv:2607.10400},
  year={2026}
}

@article{ref72,
  title={DocAtlas: Long-Document Understanding as Mutable-State Interaction},
  author={Wei, Hongchen and Wang, Yuanzhe and Liu, Bei and Yang, Yifan and Dai, Qi and Qiu, Kai and Li, Yunsheng and Chen, Dongdong and Luo, Chong and Chen, Zhenzhong and Guo, Baining},
  journal={arXiv preprint arXiv:2608.07527},
  year={2026}
}

@inproceedings{ref75,
  title={Chartqa: A benchmark for question answering about charts with visual and logical reasoning},
  author={Masry, Ahmed and Jia Qing Tan and Shafiq Joty and Enamul Hoque},
  booktitle={Findings of the association for computational linguistics: ACL 2022},
  year={2022},
  pages={2263--2279},
  month={May}
}

@inproceedings{ref76,
  title={Pix2struct: Screenshot parsing as pretraining for visual language understanding},
  author={Lee, Kenton and Mandar Joshi and Iulia Raluca Turc and Hexiang Hu and Fangyu Liu and Julian Martin Eisenschlos and Urvashi Khandelwal and Peter Shaw and Ming-Wei Chang and Kristina Toutanova},
  booktitle={International Conference on Machine Learning},
  year={2023},
  pages={18893--18912},
  month={July},
  organization={PMLR}
}

@inproceedings{ref79,
  title={CMDR: Contextual Multimodal Document Retrieval},
  author={Tanaka, Ryota and Taku Hasegawa and Kyosuke Nishida},
  booktitle={European Conference on Computer Vision},
  year={2026},
  pages={77--95},
  month={September},
  publisher={Springer Nature Switzerland},
  address={Cham}
}

@article{ref82,
  title={Vidore benchmark v2: Raising the bar for visual retrieval},
  author={Mac{\'e}, Quentin and Ant{\'o}nio Loison and Manuel Faysse},
  journal={arXiv preprint arXiv:2505.17166},
  year={2025}
}

@article{ref134,
  title={HierDoc: Hierarchical Page-to-Region Evidence Routing for Long-Document Visual Question Answering},
  author={Gu, Rongjian and Wengang Zhou and Junyu Xiong and Yonghui Wang and Bing Yin and Bei Wang and Houqiang Li},
  journal={arXiv preprint arXiv:2607.29638},
  year={2026}
}

@article{ref145,
  title={DocClaw: A Unified Agentic System for Intelligent Document Processing},
  author={Xiang, Siqi and Zhipeng Xu and Yufei Liu and Junhao Ji and Qing Liu and Zulong Chen and Zhibo Yang and Chunyan Miao and Shijian Lu},
  journal={arXiv preprint arXiv:2608.18685},
  year={2026}
}

@article{ref148,
  title={DocMemo: Dynamic Evidence Discovery via Probabilistic Memory-Guided Retrieval for Multi-Modal Document Understanding},
  author={Yao, Hanshu and Janfeng Zhong and Niu Lian and Jinpeng Wang},
  journal={arXiv preprint arXiv:2608.07067},
  year={2026}
}

@article{ref149,
  title={D2-ScaleAgent: Dual-Dimensional Scaling for Long Document Understanding},
  author={Zhang, Hao and Longrong Yang and Lunhao Duan and Ziyang Wang and Qing-Guo Chen and Shanshan Zhao},
  journal={arXiv preprint arXiv:2608.16417},
  year={2026}
}

@article{ref150,
  title={Navigating Sparse Evidence: Agentic Visual RAG via Explicit Context Selection and Consolidation},
  author={Shen, Yucheng and Lingyong Yan and Jiulong Wu and Shuaiqiang Wang and Jianmin WU and Dawei Yin and Min Cao},
  journal={arXiv preprint arXiv:2609.15800},
  year={2026}
}

@inproceedings{ref156,
  title={Neusym-rag: Hybrid neural symbolic retrieval with multiview structuring for pdf question answering},
  author={Cao, Ruisheng and Zhang, Hanchong and Huang, Tiancheng and Kang, Zhangyi and Zhang, Yuxin and Sun, Liangtai and Li, Hanqi and Miao, Yuxun and Fan, Shuai and Chen, Lu and Yu, Kai},
  booktitle={Proceedings of the 63rd Annual Meeting of the Association for Computational Linguistics (Volume 1: Long Papers)},
  year={2025},
  pages={6211--6239},
  month={July}
}

@inproceedings{ref157,
  title={AirQA: A Comprehensive QA Dataset for AI Research with Instance-Level Evaluation},
  author={Huang, Tiancheng and Cao, Ruisheng and Zhang, Yuxin and Kang, Zhangyi and Wang, Zijian and Wang, Chenrun and Luo, Yijie and Zheng, Hang and Qian, Lirong and Chen, Lu and Yu, Kai},
  booktitle={International Conference on Learning Representations},
  year={2026},
  volume={2026},
  pages={9927--9955},
  month={April}
}

@article{ref5,
  title={Spiqa: A dataset for multimodal question answering on scientific papers},
  author={Pramanick, Shraman and Rama Chellappa and Subhashini Venugopalan},
  journal={Advances in Neural Information Processing Systems},
  volume={37},
  pages={118807--118833},
  year={2024}
}

@inproceedings{ref12,
  title={Docbench: A benchmark for evaluating llm-based document reading systems},
  author={Zou, Anni and Wenhao Yu and Hongming Zhang and Kaixin Ma and Deng Cai and Zhuosheng Zhang and Hai Zhao and Dong Yu},
  booktitle={Proceedings of the 4th international workshop on knowledge-augmented methods for natural language processing},
  pages={359--373},
  year={2025}
}

@inproceedings{ref14,
  title={Longdocurl: a comprehensive multimodal long document benchmark integrating understanding, reasoning, and locating},
  author={Deng, Chao and Yuan, Jiale and Bu, Pi and Wang, Peijie and Li, Zhong-Zhi and Xu, Jian and Li, Xiao-Hui and Gao, Yuan and Song, Jun and Zheng, Bo and Liu, Cheng-Lin},
  booktitle={Proceedings of the 63rd Annual Meeting of the Association for Computational Linguistics (Volume 1: Long Papers)},
  pages={1135--1159},
  year={2025}
}

@inproceedings{ref19,
  title={DocMMIR: A Framework for Document Multi-modal Information Retrieval},
  author={Li, Zirui and Siwei Wu and Yizhi Li and Xingyu Wang and Yi Zhou and Chenghua Lin},
  booktitle={EMNLP (Findings)},
  pages={13117--13130},
  year={2025}
}

@inproceedings{ref22,
  title={Doclayllm: An efficient multi-modal extension of large language models for text-rich document understanding},
  author={Liao, Wenhui and Jiapeng Wang and Hongliang Li and Chengyu Wang and Jun Huang and Lianwen Jin},
  booktitle={2025 IEEE/CVF Conference on Computer Vision and Pattern Recognition (CVPR)},
  pages={4038--4049},
  publisher={IEEE},
  year={2025}
}

@article{ref23,
  title={DocVAL: Validated Chain-of-Thought Distillation for Grounded Document VQA},
  author={Neogi, Pinaki Prasad Guha and Ahmad Mohammadshirazi and Ser-Nam Lim and Rajiv Ramnath},
  journal={arXiv preprint arXiv:2511.22521},
  year={2025}
}

@inproceedings{ref24,
  title={Uni-DocRobust: Universal Plug-and-Play Robustness Enhancement for Multi-modal LLMs via Feature Restoration},
  author={Zhou, Yuxuan and Baole Wei and Xingjian Hu and Haowei Chen and Yu Li and Xingyue Lin and Liangcai Gao and Zhi Tang},
  booktitle={Forty-third International Conference on Machine Learning},
  year={2026}
}

@inproceedings{ref31,
  title={Towards storage-efficient visual document retrieval: An empirical study on reducing patch-level embeddings},
  author={Ma, Yubo and Li, Jinsong and Zang, Yuhang and Wu, Xiaobao and Dong, Xiaoyi and Zhang, Pan and Cao, Yuhang and Duan, Haodong and Wang, Jiaqi and Cao, Yixin and Sun, Aixin},
  booktitle={Findings of the Association for Computational Linguistics: ACL 2025},
  pages={19568--19580},
  year={2025}
}

@inproceedings{ref57,
  title={TRACE: Traversal Retrieval-Augmented Chain of Evidence for Document Understanding},
  author={He, Liqi and Zuchao Li and Hao Huang and Ping Wang},
  booktitle={Proceedings of the 64th Annual Meeting of the Association for Computational Linguistics (Volume 1: Long Papers)},
  pages={9806--9825},
  year={2026}
}

@article{ref58,
  title={XL-DocBench: Benchmarking Evidence-Grounded Extra-Long Document Understanding},
  author={Wei, Hongchen and Wang, Yuanzhe and Liu, Bei and Yang, Yifan and Dai, Qi and Ma, Ruichun and Qiu, Kai and Li, Yunsheng and Chen, Dongdong and Luo, Chong and Chen, Zhenzhong and Guo, Baining},
  journal={arXiv preprint arXiv:2608.00036},
  year={2026}
}

@inproceedings{ref63,
  title={Docslm: A small vision-language model for long multimodal document understanding},
  author={Hannan, Tanveer and Dimitrios Mallios and Parth Pathak and Faegheh Sardari and Thomas Seidl and Gedas Bertasius and Mohsen Fayyaz and Sunando Sengupta},
  booktitle={Proceedings of the IEEE/CVF Conference on Computer Vision and Pattern Recognition},
  pages={9337--9346},
  year={2026}
}

@inproceedings{ref66,
  title={ICDAR2026 Competition on Multimodal Reasoning over Documents in Multiple Domains},
  author={Llabrés, Artemis and Marc Serra Ortega and Tomàs Ockier and Samuel Ortega Cuadra and Amritpal Singh and Christos Georgakilas and Andrey Barsky and Ernest Valveny and Dimosthenis Karatzas},
  booktitle={International Conference on Document Analysis and Recognition},
  pages={336--352},
  publisher={Springer Nature Switzerland},
  address={Cham},
  year={2026}
}

@article{ref74,
  title={How to Train Your Long-Context Visual Document Model},
  author={Veselka, Austin},
  journal={arXiv preprint arXiv:2602.15257},
  year={2026}
}

@inproceedings{ref78,
  title={Multimodal arxiv: A dataset for improving scientific comprehension of large vision-language models},
  author={Li, Lei and Yuqi Wang and Runxin Xu and Peiyi Wang and Xiachong Feng and Lingpeng Kong and Qi Liu},
  booktitle={Proceedings of the 62nd Annual Meeting of the Association for Computational Linguistics (Volume 1: Long Papers)},
  pages={14369--14387},
  year={2024}
}

@inproceedings{ref80,
  title={Hybrid-vector retrieval for visually rich documents: Combining single-vector efficiency and multi-vector accuracy},
  author={Kim, Juyeon and Geon Lee and Dongwon Choi and Taeuk Kim and Kijung Shin},
  booktitle={Findings of the Association for Computational Linguistics: ACL 2026},
  pages={1073--1089},
  year={2026}
}

@inproceedings{ref93,
  title={mplug-docowl 1.5: Unified structure learning for ocr-free document understanding},
  author={Hu, Anwen and Haiyang Xu and Jiabo Ye and Ming Yan and Liang Zhang and Bo Zhang and Ji Zhang and Qin Jin and Fei Huang and Jingren Zhou},
  booktitle={Findings of the Association for Computational Linguistics: EMNLP 2024},
  pages={3096--3120},
  year={2024}
}

@article{ref98,
  title={VinQA: Visual Elements Interleaved Long-form Answer Generation for Real-World Multimodal Document QA},
  author={Jang, Young Rok and Hyesoo Kong and Kyunghwan An and Jae Sub Huh and Gyeonghun Kim and Stanley Jungkyu Choi},
  journal={arXiv preprint arXiv:2606.16092},
  year={2026}
}

@article{ref100,
  title={Uda: A benchmark suite for retrieval augmented generation in real-world document analysis},
  author={Hui, Yulong and Yao Lu and Huanchen Zhang},
  journal={Advances in Neural Information Processing Systems},
  volume={37},
  pages={67200--67217},
  year={2024}
}

@article{ref105,
  title={Benchmarking retrieval-augmented multimomal generation for document question answering},
  author={Dong, Kuicai and Chang, Yujing and Huang, Shijie and Wang, Yasheng and Tang, Ruiming and Liu, Yong},
  journal={Advances in Neural Information Processing Systems},
  volume={38},
  year={2026}
}

@inproceedings{ref106,
  title={DocRetriever: A Plug-and-Play Framework for Multimodal Document Retrieval with Comprehensive Benchmark},
  author={Hu, Ruofan and Zhu, Menghui and Zhu, Jieming and Chen, Bo and Xu, Shengyang and Hong, Minjie and Yang, Xiaoda and Zhou, Sashuai and Tang, Li and Jin, Tao and Zhao, Zhou},
  booktitle={Proceedings of the 32nd ACM SIGKDD Conference on Knowledge Discovery and Data Mining V. 2},
  pages={1780--1791},
  year={2026}
}

@inproceedings{ref109,
  title={Finragbench-v: A benchmark for multimodal rag with visual citation in the financial domain},
  author={Zhao, Suifeng and Zhuoran Jin and Sujian Li and Jun Gao},
  booktitle={Proceedings of the 2025 Conference on Empirical Methods in Natural Language Processing},
  pages={4215--4249},
  year={2025}
}

@article{ref163,
  title={MemoryDocDataSet: A Benchmark for Joint Conversational Memory and Long Document Reasoning},
  author={Xie, Qiyang and Jialun Wu and Xinjie He and Su Liu and Shuai Xiao and Zhiyuan Lin and Weikai Zhou},
  journal={arXiv preprint arXiv:2606.04442},
  year={2026}
}

@article{ref164,
  title={Memskill: Learning and evolving memory skills for self-evolving agents},
  author={Zhang, Haozhen and Quanyu Long and Jianzhu Bao and Tao Feng and Weizhi Zhang and Haodong Yue and Wenya Wang},
  journal={arXiv preprint arXiv:2602.02474},
  year={2026}
}

@article{ref165,
  title={Recursive self-improvement of AI research agents},
  author={Srikanth, Dhruv and Bingchen Zhao and Dixing Xu and Yuxiang Wu and Zhengyao Jiang},
  journal={arXiv preprint arXiv:2609.26457},
  year={2026}
}

@article{ref110,
  title={Look as you think: Unifying reasoning and visual evidence attribution for verifiable document RAG via reinforcement learning},
  author={Liu, Shuochen and Pengfei Luo and Chao Zhang and Yuhao Chen and Haotian Zhang and Qi Liu and Xin Kou and Tong Xu and Enhong Chen},
  journal={Proceedings of the AAAI Conference on Artificial Intelligence},
  volume={40},
  number={38},
  pages={32159--32167},
  year={2026}
}

@article{ref111,
  title={Chain of Evidence: Pixel-Level Visual Attribution for Iterative Retrieval-Augmented Generation},
  author={Liu, Peiyang and Ziqiang Cui and Xi Wang and Di Liang and Wei Ye},
  journal={arXiv preprint arXiv:2605.01284},
  year={2026}
}

@article{ref112,
  title={Evidence Attribution in Visual Document Understanding without Coordinates or Region Labels},
  author={Liu, Zhuchenyang and Yao Zhang and Yu Xiao},
  journal={arXiv preprint arXiv:2607.24651},
  year={2026}
}

@inproceedings{ref119,
  title={M3Grounder: Mask-Based Multi-Span and Multi-Granular Grounding for Document QA},
  author={Venna, Venkata Kesav and Sai Madhusudan Gunda and Jyothi Swaroopa Jinka and Hrithik Sagar Rachakonda and Anirudh Srinivasan and Ravi Kiran Sarvadevabhatla},
  booktitle={Proceedings of the IEEE/CVF Conference on Computer Vision and Pattern Recognition},
  pages={23685--23695},
  year={2026}
}

@inproceedings{ref120,
  title={DoCoG: Mask-Based Multi-type Grounded Chain-of-Thought for Document QA},
  author={Gunda, Sai Madhusudan and Jyothi Swaroopa Jinka and Hrithik Sagar Rachakonda and Aryan Jain and Venkata Kesav Venna and Anirudh Srinivasan and Ravi Kiran Sarvadevabhatla},
  booktitle={European Conference on Computer Vision},
  pages={308--327},
  publisher={Springer Nature Switzerland},
  address={Cham},
  year={2026}
}

@article{ref122,
  title={DocPC: Document-Level Visual Retrieval via Representative Page Composition},
  author={You, Chengsong and Zhou, Junwei and Cao, Xiaoyu and Wang, Weiyao and Xu, Yiwei and Zhao, Ziyan and Sun, Zhen and Zhu, Qicheng and Fu, Xuanyi and Chen, Yufan and others},
  journal={arXiv preprint arXiv:2608.25434},
  year={2026}
}

@inproceedings{ref124,
  title={LayoutLLM: Large language model instruction tuning for visually rich document understanding},
  author={Fujitake, Masato},
  booktitle={Proceedings of the 2024 Joint International Conference on Computational Linguistics, Language Resources and Evaluation (LREC-COLING 2024)},
  pages={10219--10224},
  year={2024}
}

@article{ref130,
  title={LFRAG: Layout-oriented Fine-grained Retrieval-Augmented Generation on Multimodal Document Understanding},
  author={Zhu, Yifan and Yu Mi and Yue Lu and Yanchu Guan and Zhixuan Chu},
  journal={arXiv preprint arXiv:2605.22829},
  year={2026}
}

@article{ref133,
  title={Are we on the right way to assess document retrieval-augmented generation?},
  author={Shen, Wenxuan and Mingjia Wang and Yaochen Wang and Dongping Chen and Junjie Yang and Yao Wan and Weiwei Lin},
  journal={Proceedings of the AAAI Conference on Artificial Intelligence},
  volume={40},
  number={39},
  pages={32929--32937},
  year={2026}
}

@inproceedings{ref136,
  title={PaperScope: A Multi-Modal Multi-Document Benchmark for Agentic Deep Research Across Massive Scientific Papers},
  author={Xiong, Lei and Huaying Yuan and Zheng Liu and Zhao Cao and Zhicheng Dou},
  booktitle={Findings of the Association for Computational Linguistics: ACL 2026},
  pages={8015--8040},
  year={2026}
}

@inproceedings{ref139,
  title={Docopilot: Improving multimodal models for document-level understanding},
  author={Duan, Yuchen and Chen, Zhe and Hu, Yusong and Wang, Weiyun and Ye, Shenglong and Shi, Botian and Lu, Lewei and Hou, Qibin and Lu, Tong and Li, Hongsheng and Dai, Jifeng and Wang, Wenhai},
  booktitle={2025 IEEE/CVF Conference on Computer Vision and Pattern Recognition (CVPR)},
  pages={4026--4037},
  publisher={IEEE},
  year={2025}
}

@misc{ref140,
  title={RegionDoc-R1: Reinforcing Semantic Layout-Aware Learning for Document Understanding},
  author={Wang, Chao and Hehe Fan and Huichen Yang and Zhengdong Hu and Sarvnaz Karimi and Lina Yao and Yi Yang},
  year={2026}
}

@article{ref142,
  title={Wukong: A large multimodal model for efficient long pdf reading with end-to-end sparse sampling},
  author={Xie, Xudong and Hao Yan and Liang Yin and Yang Liu and Jing Ding and Minghui Liao and Yuliang Liu and Wei Chen and Xiang Bai},
  journal={International Journal of Computer Vision},
  volume={134},
  number={7},
  pages={324},
  year={2026}
}

\clearpage
\appendix

\section{Benchmark Catalog}
\label{app:benchmarks}

Beyond the representative comparison in Table~\ref{tab:benchmarks}, Table~\ref{tab:benchmark-catalog} catalogs all 55 benchmark records covered in this survey, retaining their scale, question types, and evaluation capabilities.
Benchmarks are grouped by their primary evaluation goal into retrieval-oriented and answer-quality-oriented benchmarks, with the latter further grouped by input page count.
Benchmark selection should distinguish input scale, evidence span, and the stage at which processing begins.
Longer inputs do not necessarily require cross-page evidence combination: MP-DocVQA identifies single-page evidence among multiple candidate pages \citep{tito2023hierarchical}, whereas SlideVQA includes questions requiring evidence across pages \citep{tanaka2023slidevqa}.
An open candidate corpus and a given document reflect different task settings, so page counts alone do not establish relative difficulty.
Benchmarks such as SPIQA provide pre-extracted elements \citep{ref5}; their results primarily measure understanding and evidence selection over the supplied material, rather than full-page parsing.
Evidence annotation granularity further determines what can be verified: document or page retrieval, region localization, and answer quality should be interpreted separately.

MMLongBench-Doc and M-LongDoc assess long-input processing \citep{ma2024mmlongbench,ref16}; SlideVQA and CMDR-Bench probe cross-page evidence requirements \citep{tanaka2023slidevqa,ref79}; and MMDocIR and VISA support fine-grained acquisition and source verification \citep{ref17,ref13}.
DVisRAG evaluates performance under visual degradation \citep{ref89}, while DocScope analyzes answering bottlenecks under different evidence conditions \citep{ref113}.
Cost comparisons should specify input conditions and budgets and account for parsing, indexing, and online inference, rather than assessing system capabilities through answer scores alone.

\section{Methods without Explicit Retrieval}
\label{app:direct-reading}

Table~\ref{tab:direct-methods-full} extends the direct-reading comparison in Table~\ref{tab:direct-methods}.
These methods organize reading within the supplied content without explicit evidence retrieval in the answering pipeline, and are grouped into single-page and multipage methods according to their primary design and evaluation settings.
Multipage methods make different trade-offs among information organization, input scale, and evidence use.
Hi-VT5 organizes multipage inputs through hierarchical representations \citep{tito2023hierarchical}, GRAM uses document tokens for cross-page communication \citep{ref117}, and mPLUG-DocOwl2 compresses high-resolution visual features to control input scale \citep{ref26}.
M3D and DocR1 strengthen the link between evidence identification and answering through evidence-and-answer supervision and evidence-page rewards, respectively \citep{tanaka2023slidevqa,ref27}.
These designs act on representations, fusion, or learning objectives; Table~\ref{tab:direct-methods-full} compares external extraction dependence, generation organization, backbone models, added training, and the dimensions actually evaluated.

Direct reading allocates resources among input capacity, detail fidelity, and cross-page relations.
Compression reduces representation size but may lose critical text or visual details.
More input pages broaden coverage and higher resolution improves detail visibility, but both increase reading cost.
Comparisons should therefore consider answer quality, evidence selection, and cost together, distinguishing model capabilities from input configurations.
Direct reading and explicit retrieval are complementary: hierarchical fusion and compression support multipage reading after retrieval, while retrieval further controls the evidence passed to generation.

\begin{figure*}[t]
  \centering
  \includegraphics[width=0.95\textwidth]{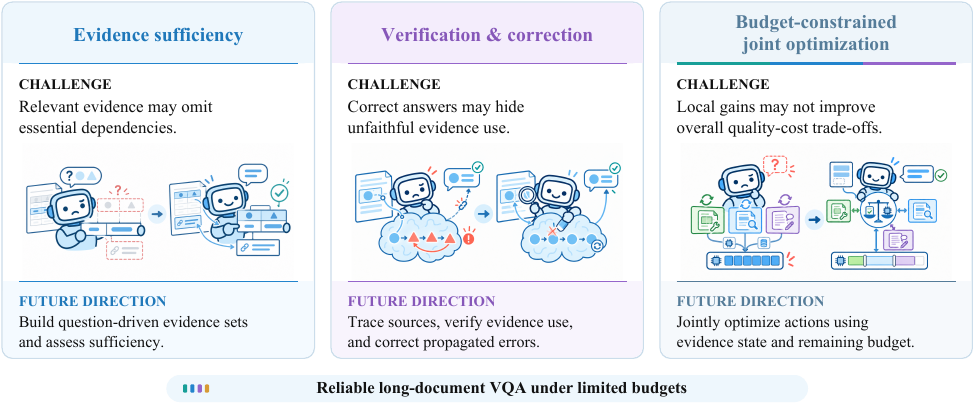}
  \caption{Challenges and future directions for answering individual questions in long-document VQA. The three panels correspond to Secs.~\ref{app:evidence-sufficiency}--\ref{app:budget-decisions}, covering evidence sufficiency, verifiable and correctable understanding, and budget-aware coordination toward reliable answers under limited budgets.}
  \label{fig:challenges-directions}
  \vspace{-10pt}
\end{figure*}

\section{Methods with Explicit Retrieval}
\label{app:explicit-retrieval}

Table~\ref{tab:explicit-methods-full} catalogs 110 explicit-retrieval methods, extending the representative comparison in Table~\ref{tab:explicit-methods}.
For retrieval-only methods, the comparison covers evidence representations, matching, and returned results; generation attributes do not apply.
For QA methods, it additionally records how evidence reaches the generator, agent organization, reasoning feedback, and added training.
Parsing, retrieval, generation, and evaluation attributes should be read together: the same visual retriever can support different generation pipelines, and neither structural parsing nor multi-agent organization alone determines whether acquisition is iterative.

Retriever matching representations should be distinguished from generator inputs.
\emph{Modality} denotes the evidence-side modalities used to assess relevance, whereas \emph{Granularity} denotes the evidence units actually received by the generator; retrieval-only methods are labeled by their returned units.
When retrieval hits and generation inputs differ in modality or granularity, \emph{Mechanism} records the mapping with an arrow.
For example, ViDoRAG maps retrieved text chunks to their source page images \citep{ref18}, allowing text matching to support visual page reading.
\emph{Iterative} acquisition introduces evidence beyond the current set in response to generation or control feedback; graph traversal without such feedback, fixed multistage retrieval, and reranking or refinement of existing evidence do not qualify.
\emph{Reasoning} instead describes whether generation involves a feedback loop.
Thus, One-shot acquisition can coexist with Multi-step reasoning, whereas a single CoT or a fixed multi-agent pipeline is classified as Single-pass.

Superscript $t$ in \emph{Mechanism} marks added retrieval, scoring, or page-selection training.
Generation-side \emph{Learning} records added training of answering, control, or tool components; it does not imply that all modules are updated together.
For example, MM-Doc-R1 and DocDancer train only the language agent while freezing the visual Read tool \citep{ref51,ref52}; AgenticOCR trains only the region tool agent while freezing the final answerer \citep{ref104}; Doc-REFRAG applies RL only to its token-recovery module \citep{ref59}; and RobustVisRAG performs feature-alignment training only on the generation-side visual encoder while freezing the language backbone \citep{ref89}.

\section{Challenges and Research Opportunities}
\label{app:research-opportunities}

Building on Sec.~\ref{sec:future-directions}, this appendix discusses challenges and research opportunities at two levels: improving capabilities within the current task and extending its boundaries.
Secs.~\ref{app:evidence-sufficiency}--\ref{app:budget-decisions} examine evidence sufficiency, verifiable and correctable understanding, and budget-aware coordination for individual questions, drawing on existing advances and limitations to identify future directions and concrete research opportunities (Fig.~\ref{fig:challenges-directions}).
Sec.~\ref{app:continual-understanding} extends the independent-question setting through memory reuse across questions and strategy improvement across tasks, broadening the perspective to sustained interaction and accumulated experience.

\subsection{Evidence Sufficiency}
\label{app:evidence-sufficiency}

Task-oriented evidence sufficiency moves beyond local relevance to the support required for an answer, asking: \emph{Is the current evidence sufficient, and what is missing?}
CMDR-Embed incorporates cross-page context into retrieval representations \citep{ref79}, while AgenticOCR narrows page reading to question-relevant regions \citep{ref104}.
Yet contextual association and fine-grained extraction can still omit essential dependencies.
AgenticOCR explicitly notes that extracting only a table row or cell without its header and surrounding context can leave evidence incomplete.
Finding relevant content therefore does not ensure sufficient support for answering.

\textbf{Future direction.}
Move from matching individual evidence units by relevance to question-driven evidence set construction and sufficiency assessment.
This requires identifying the facts needed for an answer, their dependencies, and the remaining gaps, providing testable grounds for adaptive acquisition and stopping.

\textbf{Research opportunities.}
\emph{(a) Model evidence requirements and dependencies.}
Organize evidence around the facts, comparisons, and calculation conditions required by the question, distinguishing context used for localization from evidence that directly supports the answer.
\emph{(b) Select complementary evidence jointly.}
Treat tables continued across pages, headers and values, or figures and their annotations as interdependent evidence combinations; balance coverage against redundancy and avoid fragmentation that breaks essential semantic dependencies.
\emph{(c) Learn testable sufficiency judgments.}
Construct examples with complete evidence, partially missing evidence, and highly relevant but insufficient evidence to test whether models can identify gaps and translate them into explicit acquisition needs.

\subsection{Verifiable and Correctable Understanding}
\label{app:verifiable-understanding}

Verifiable and correctable document understanding moves beyond correct answers to trustworthy evidence chains, asking: \emph{Is acquired information used faithfully, and can errors be corrected?}
Methods such as DocMemo explicitly maintain evidence states across rounds \citep{ref148}, while DocScope verifies pages, regions, facts, and answers \citep{ref113}.
DocScope's experiments show that correct answers can coexist with incomplete evidence chains.
This raises a further question: can parsing or summarization errors enter memory, affect subsequent acquisition, and become repeatedly reinforced?

\begin{samepage}
\textbf{Future direction.}
Connect source tracing, process verification, and error correction to maintain traceable, correctable evidence states across rounds, and evaluate whether systems can detect, locate, and correct errors.\par
\end{samepage}

\textbf{Research opportunities.}
\emph{(a) Build revisable evidence memory.}
Retain source pages, regions, and dependencies for intermediate facts and summaries; when misreadings or conflicts emerge, reread the sources, retract erroneous records, and update dependent judgments.
\emph{(b) Test whether evidence supports the answer.}
Remove or replace key evidence under controlled conditions to test whether models detect missing support or changed facts, while checking stability under changes to irrelevant or redundant evidence.
Distinguish citation accuracy, accurate fact reading, and evidential support for the answer.
\emph{(c) Connect diagnosis to correction.}
Under cross-page dependencies, distractors, visual degradation, and unanswerable conditions, evaluate whether systems choose to reparse, acquire more evidence, revise the answer, or abstain, and measure their ability to recover from errors.

\subsection{Decision Making under Budget Constraints}
\label{app:budget-decisions}

Budget-aware coordination moves from local optimization to end-to-end quality--cost trade-offs, asking: \emph{Where should limited computation be spent?}
Visual compression, tool-action learning in VRAG-RL \citep{wang2026vragrl}, on-demand parsing in AgenticOCR \citep{ref104}, and memory-guided retrieval in DocMemo \citep{ref148} provide several forms of dynamic processing.
However, savings in individual operations may be partly offset by auxiliary model calls, multiple interaction rounds, and state maintenance.
Their quality gains and computational costs must therefore be assessed across the full pipeline.

\textbf{Future direction.}
Building on dynamic processing and cross-component interactions, use evidence states and remaining budgets to jointly decide whether to search further, inspect local details, reparse, verify a judgment, or finish answering.
Learn evidence acquisition and use jointly under a quality--cost objective.

\textbf{Research opportunities.}
\emph{(a) Allocate budgets dynamically across components.}
Adapt parsing scope, visual resolution, retrieval depth, and reasoning rounds to question difficulty and current evidence gaps; determine when direct reading is preferable and when retrieval or tools are worthwhile.
\emph{(b) Learn the marginal value of actions.}
Extend action learning by comparing how much an additional retrieval, local rereading, or verification step improves evidence coverage and answer quality, and incorporate cost into action selection and stopping policies.
\emph{(c) Test generalization across budgets and settings.}
Compare quality--cost curves across document lengths, degrees of evidence dispersion, and computational budgets.
Include preprocessing, indexing, auxiliary models, and online inference to assess whether policies adapt to practical resource constraints.

\begin{figure*}[t]
  \centering
  \includegraphics[width=0.95\textwidth]{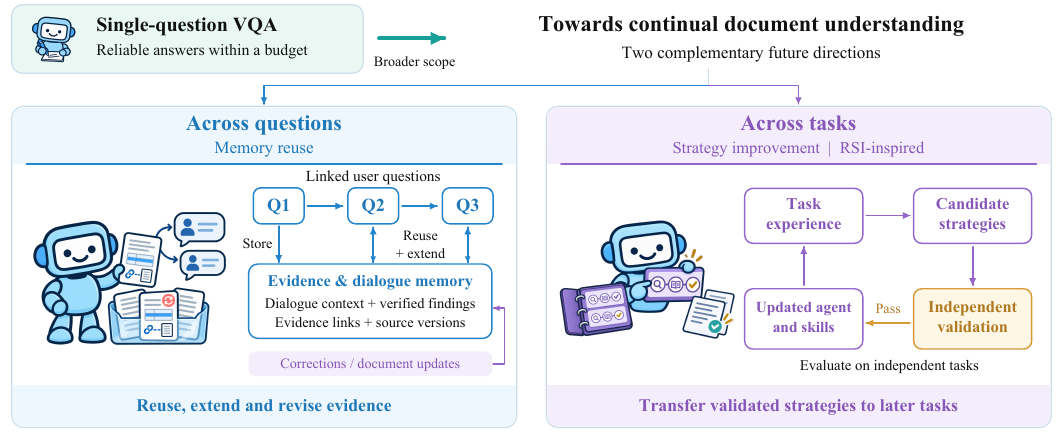}
  \caption{Two complementary directions from individual question answering toward continual document understanding. Across questions, systems reuse, supplement, and revise evidence and interaction memory. Across tasks, they propose strategies from experience, validate them on independent tasks before updating agents and skills, and continue accumulating experience.}
  \label{fig:continual-understanding}
  \vspace{-10pt}
\end{figure*}

\subsection{Towards Continual Document Understanding}
\label{app:continual-understanding}

The preceding three directions focus on answering individual questions.
Building on them, this survey considers how to extend document VQA beyond independent questions.
Practical document use requires systems to solve the current question, exploit connections across successive interactions, and improve subsequent processing strategies through experience across tasks.
We therefore identify two related extensions---memory reuse across questions and continual, experience-driven strategy improvement---that shift the perspective from individual answers to sustained interaction and accumulated capabilities (Fig.~\ref{fig:continual-understanding}).

\textbf{Memory reuse across questions.}
Multiturn document VQA over successive user questions must combine document content with interaction history to resolve cross-turn references, carry forward comparison conditions, incorporate user corrections, and reuse evidence that remains applicable.
MemoryDocDataSet explores the combination of conversational memory and reasoning over long text documents \citep{ref163}.
For visual documents, future systems could link dialogue to source pages and regions, retaining provenance, versions, and verified results to reuse earlier localization and reading while acquiring missing support for new questions.
For example, a follow-up about another metric in the same table can reuse the table's location and measurement conventions.
The key is to assess whether historical information remains applicable and revise affected memories after user corrections or document updates, reducing repeated processing and error propagation.

\textbf{Continual, experience-driven strategy improvement.}
Beyond reusing specific evidence, systems could accumulate transferable processing strategies from experience across tasks.
Because document types differ in structure and evidence dependencies, systems need to learn which retrieval, reading, and verification procedures suit which conditions.
MemSkill revises memory skills through difficult cases \citep{ref164}, while AIDE$^2$ enables updated agents to continue participating in their own improvement \citep{ref165}, offering insights from general-purpose agents.
Inspired by recursive self-improvement (RSI), future systems could propose strategies from document-QA successes, failures, and user feedback, validate them on independent tasks before updating a skill library or control procedure, and let the updated system participate in subsequent strategy revision.
For example, experience with financial reports could yield a procedure that checks units and reporting periods before comparing values.
The central challenge is to establish applicability conditions and feedback grounds that support strategy transfer to subsequent documents.

\clearpage

\begingroup
\newcommand{\TableFourYes}{\MethodYes}
\newcommand{\TableFourNo}{\MethodNo}

\makeatletter
\setlength{\@fptop}{0pt}
\setlength{\@dblfptop}{0pt}
\makeatother

\begin{table*}[!t]
\captionsetup{position=top,skip=3pt}
\caption{Complete benchmark catalog: 55 records covering dataset size, question types, evidence annotations, and evaluation capabilities. Ordering and first-release dates follow Table~\ref{tab:explicit-methods}. -- denotes unreported or inapplicable information.}
\label{tab:benchmark-catalog}
\centering
\begingroup
\fontsize{7.8}{8.8}\selectfont
\setlength{\tabcolsep}{1.3pt}
\renewcommand{\arraystretch}{0.96}
\setlength{\aboverulesep}{1.0pt}
\setlength{\belowrulesep}{1.0pt}
\resizebox{\textwidth}{!}{%
\begin{tabular}{@{}l*{13}{c}@{}}
\toprule
\multicolumn{3}{c}{\textbf{Benchmark}} & \multicolumn{4}{c}{\textbf{Data Size}} & \multicolumn{4}{c}{\textbf{Question Type}} & \multicolumn{3}{c}{\textbf{Evidence and Evaluation}} \\
\cmidrule(lr){1-3}\cmidrule(lr){4-7}\cmidrule(lr){8-11}\cmidrule(l){12-14}
\textbf{Name} & \textbf{Venue} & \textbf{Domain} & \textbf{\#Docs} & \textbf{\#Pages} & \textbf{\#Tokens} & \textbf{\#QA} & \textbf{MD} & \textbf{MP} & \textbf{SP} & \textbf{Unans.} & \textbf{Source} & \textbf{Gran.} & \textbf{Eval.} \\
\midrule
\multicolumn{14}{@{}l@{}}{\textit{Retrieval-focused benchmarks}} \\
\addlinespace[1.0pt]
Wiki-SS~\citeyearpar{ma2024unifying}~\MethodDate{24.06} & EMNLP\allowbreak\textquoteright{}24 & Open & -- & 1.27M & -- & 52.7k & \TableFourNo{} & \TableFourNo{} & \TableFourYes{} & \TableFourNo{} & TXT & Page & Ans. Ret. \\
SlideVQA-open~\citeyearpar{ma2024unifying}~\MethodDate{24.06} & EMNLP\allowbreak\textquoteright{}24 & Open & -- & 50.7k & -- & 12.4k & \TableFourNo{} & \TableFourYes{} & \TableFourYes{} & \TableFourNo{} & TXT/\allowbreak{}L/\allowbreak{}C/\allowbreak{}TAB/\allowbreak{}I & Page & Ret. \\
ViDoRe~\citeyearpar{ref7}~\MethodDate{24.06} & ICLR\allowbreak\textquoteright{}25 & Open & -- & 1.0k & -- & 3.81k & \TableFourNo{} & \TableFourNo{} & \TableFourYes{} & \TableFourNo{} & TXT/\allowbreak{}L/\allowbreak{}C/\allowbreak{}TAB/\allowbreak{}I & Page & Ans.\textsuperscript{a} Ret. \\
REAL-MM-RAG~\citeyearpar{ref88}~\MethodDate{25.02} & ACL\allowbreak\textquoteright{}25 & Open & 163 & 2.1k & -- & 4.6k & \TableFourNo{} & \TableFourNo{} & \TableFourYes{} & \TableFourNo{} & TXT/\allowbreak{}C/\allowbreak{}TAB/\allowbreak{}I & Page & Ans. Ret. \\
ViDoRe V2~\citeyearpar{ref82}~\MethodDate{25.03} & arXiv\allowbreak\textquoteright{}25 & Open & 62 & 1.1k & -- & 327 & \TableFourYes{} & \TableFourYes{} & \TableFourYes{} & \TableFourNo{} & TXT/\allowbreak{}L/\allowbreak{}C/\allowbreak{}TAB/\allowbreak{}I & Page & Ret. \\
MMDocIR~\citeyearpar{ref17}~\MethodDate{25.01} & EMNLP\allowbreak\textquoteright{}25 & Closed & 313 & 65.1 & -- & 1.66k & \TableFourNo{} & \TableFourYes{}(18.9\%) & \TableFourYes{}(81.1\%) & \TableFourNo{} & TXT/\allowbreak{}L/\allowbreak{}C/\allowbreak{}TAB/\allowbreak{}I & Page, Elem. & Ret. Loc. \\
DocMMIR~\citeyearpar{ref19}~\MethodDate{25.05} & EMNLP\allowbreak\textquoteright{}25\textsuperscript{F} & Open & 488.5k & -- & 537.4 & 488.5k & -- & -- & -- & -- & TXT/\allowbreak{}C/\allowbreak{}TAB/\allowbreak{}I & Doc. & Ret.\textsuperscript{a} \\
ViMDoc~\citeyearpar{ref80}~\MethodDate{25.10} & ACL\allowbreak\textquoteright{}26\textsuperscript{F} & Open & 1.4k & 76.3k & -- & 10.9k & -- & \TableFourYes{} & \TableFourYes{} & -- & TXT/\allowbreak{}L/\allowbreak{}C/\allowbreak{}TAB/\allowbreak{}I & Page & Ret. \\
MultiDocR~\citeyearpar{ref106}~\MethodDate{26.05} & KDD\allowbreak\textquoteright{}26 & Open & 313 & -- & -- & 2.58k & -- & \TableFourYes{}(43.8\%) & \TableFourYes{}(56.2\%) & \TableFourNo{} & TXT/\allowbreak{}C/\allowbreak{}TAB/\allowbreak{}I & Page & Ans. Ret. \\
CMDR-Bench~\citeyearpar{ref79}~\MethodDate{26.07} & ECCV\allowbreak\textquoteright{}26 & Closed & 255 & 183.5 & -- & 800 & \TableFourNo{} & \TableFourYes{} & -- & \TableFourNo{} & TXT/\allowbreak{}L/\allowbreak{}C/\allowbreak{}TAB/\allowbreak{}I & Page & Ret. \\
DocViRe~\citeyearpar{ref122}~\MethodDate{26.08} & arXiv\allowbreak\textquoteright{}26 & Open & 8.29k & -- & -- & 13.1k & -- & -- & -- & -- & TXT/\allowbreak{}C/\allowbreak{}TAB/\allowbreak{}I & Doc. & Ret.\textsuperscript{a} \\
\midrule
\multicolumn{14}{@{}l@{}}{\textit{Answer-quality benchmarks: \#Pages (avg.) = 1.0, \#Pages (avg.) \textgreater{} 1.0 or unreported}} \\
\addlinespace[1.0pt]
DocVQA~\citeyearpar{mathew2021docvqa}~\MethodDate{20.07} & WACV\allowbreak\textquoteright{}21 & Closed & 12.8k & 1.0 & 151.5 & 50k & \TableFourNo{} & \TableFourNo{} & \TableFourYes{} & \TableFourNo{} & TXT/\allowbreak{}L/\allowbreak{}C/\allowbreak{}TAB/\allowbreak{}I & -- & Ans. \\
InfographicVQA~\citeyearpar{ref62}~\MethodDate{21.04} & WACV\allowbreak\textquoteright{}22 & Closed & 5.5k & 1.0 & 217.9 & 30k & \TableFourNo{} & \TableFourNo{} & \TableFourYes{} & \TableFourNo{} & TXT/\allowbreak{}L/\allowbreak{}C/\allowbreak{}TAB/\allowbreak{}I & -- & Ans. \\
ChartQA~\citeyearpar{ref75}~\MethodDate{22.03} & ACL\allowbreak\textquoteright{}22\textsuperscript{F} & Closed & 20.9k & 1.0 & -- & 32.7k & \TableFourNo{} & \TableFourNo{} & \TableFourYes{} & \TableFourNo{} & C & -- & Ans. \\
ArXivQA*~\citeyearpar{ref78}~\MethodDate{24.03} & ACL\allowbreak\textquoteright{}24 & Closed & 16.6k & 1.0 & -- & 100k & \TableFourNo{} & \TableFourNo{} & \TableFourYes{} & \TableFourNo{} & C/\allowbreak{}TAB/\allowbreak{}I & -- & Ans. \\
DOGR-Bench~\citeyearpar{ref15}~\MethodDate{24.11} & ICCV\allowbreak\textquoteright{}25 & Closed & -- & 1.0 & -- & 3.6k & \TableFourNo{} & \TableFourNo{} & \TableFourYes{} & \TableFourNo{} & TXT/\allowbreak{}L/\allowbreak{}C/\allowbreak{}I & Elem. & Ans. Loc. \\
GroundingDocQA-Bench~\citeyearpar{ref119}~\MethodDate{26.06} & CVPR\allowbreak\textquoteright{}26 & Closed & 2.5k & 1.0 & -- & 5.0k & \TableFourNo{} & \TableFourNo{} & \TableFourYes{} & \TableFourNo{} & TXT/\allowbreak{}L/\allowbreak{}C/\allowbreak{}TAB/\allowbreak{}I & Elem. & Ans. Loc.\textsuperscript{a} \\
DoCoG-Bench~\citeyearpar{ref120}~\MethodDate{26.09} & ECCV\allowbreak\textquoteright{}26 & Closed & 3.0k & 1.0 & -- & 10.0k & \TableFourNo{} & \TableFourNo{} & \TableFourYes{} & \TableFourNo{} & TXT/\allowbreak{}L/\allowbreak{}C/\allowbreak{}TAB/\allowbreak{}I & Elem. & Ans. Loc.\textsuperscript{a} \\
TAT-DQA~\citeyearpar{ref86}~\MethodDate{22.07} & MM\allowbreak\textquoteright{}22 & Closed & 2.8k & 1.1 & -- & 16.6k & \TableFourNo{} & \TableFourYes{} & \TableFourYes{} & \TableFourNo{} & TXT/\allowbreak{}L/\allowbreak{}TAB & -- & Ans. \\
SlideVQA~\citeyearpar{tanaka2023slidevqa}~\MethodDate{23.01} & AAAI\allowbreak\textquoteright{}23 & Closed & 2,619 & 20.0 & 1.5k & 14.5k & \TableFourNo{} & \TableFourYes{}(13.9\%) & \TableFourYes{}(86.1\%) & \TableFourNo{} & TXT/\allowbreak{}L/\allowbreak{}C/\allowbreak{}TAB/\allowbreak{}I & Page & Ans. Ret. \\
MP-DocVQA~\citeyearpar{tito2023hierarchical}~\MethodDate{22.12} & PR\allowbreak\textquoteright{}23 & Closed & 5,928 & 8.3 & 2.0k & 46.2k & \TableFourNo{} & \TableFourNo{} & \TableFourYes{} & \TableFourNo{} & TXT/\allowbreak{}L/\allowbreak{}C/\allowbreak{}TAB/\allowbreak{}I & Page & Ans. Ret. \\
DUDE~\citeyearpar{ref3}~\MethodDate{23.05} & ICCV\allowbreak\textquoteright{}23 & Closed & 5,019 & 5.72 & 1.8k & 41.5k & \TableFourNo{} & \TableFourYes{}(2.1\%) & \TableFourYes{}(85.2\%) & \TableFourYes{}(12.7\%) & TXT/\allowbreak{}L/\allowbreak{}C/\allowbreak{}TAB/\allowbreak{}I & -- & Ans. \\
UDA~\citeyearpar{ref100}~\MethodDate{24.06} & NIPS\allowbreak\textquoteright{}24\textsuperscript{D} & Closed & 2.97k & -- & -- & 29.6k & \TableFourNo{} & -- & \TableFourYes{} & -- & TXT/\allowbreak{}TAB & Elem. & Ans. Ret.\textsuperscript{a} \\
MMLongBench-Doc~\citeyearpar{ma2024mmlongbench}~\MethodDate{24.07} & NIPS\allowbreak\textquoteright{}24\textsuperscript{D} & Closed & 135 & 47.5 & 21.2k & 1.1k & \TableFourNo{} & \TableFourYes{}(33.7\%) & \TableFourYes{}(45.7\%) & \TableFourYes{}(20.6\%) & TXT/\allowbreak{}L/\allowbreak{}C/\allowbreak{}TAB/\allowbreak{}I & Page & Ans. Ret. \\
SPIQA*~\citeyearpar{ref5}~\MethodDate{24.07} & NIPS\allowbreak\textquoteright{}24\textsuperscript{D} & Closed & 25.5k & -- & -- & 266k & \TableFourNo{} & -- & -- & \TableFourNo{} & TXT/\allowbreak{}C/\allowbreak{}TAB/\allowbreak{}I & Elem. & Ans. Ret.\textsuperscript{a} \\
VisRAG~\citeyearpar{yu2025visrag}~\MethodDate{24.10} & ICLR\allowbreak\textquoteright{}25 & Open & -- & 4.5k & -- & 3.6k & \TableFourNo{} & \TableFourYes{} & \TableFourYes{} & \TableFourNo{} & TXT/\allowbreak{}L/\allowbreak{}C/\allowbreak{}TAB/\allowbreak{}I & Page & Ans. Ret. \\
VisR-Bench~\citeyearpar{chen2024svrag}~\MethodDate{24.11} & ICLR\allowbreak\textquoteright{}25 & Closed & 226 & 13.7 & -- & 0.5k & \TableFourNo{} & \TableFourNo{} & \TableFourYes{} & \TableFourNo{} & TXT/\allowbreak{}C/\allowbreak{}TAB/\allowbreak{}I & Page & Ans. Ret. \\
PaperPDF~\citeyearpar{ref142}~\MethodDate{24.10} & IJCV\allowbreak\textquoteright{}26 & Closed & 70k & 23 & 10.2k & 1.1M & \TableFourNo{} & \TableFourYes{} & \TableFourYes{} & \TableFourNo{} & TXT/\allowbreak{}C/\allowbreak{}TAB/\allowbreak{}I & Elem. & Ans. Ret.\textsuperscript{a} \\
VisDoMBench~\citeyearpar{ref10}~\MethodDate{24.12} & NAACL\allowbreak\textquoteright{}25 & Open & 1,277 & 128.7 & -- & 2.3k & \TableFourNo{} & \TableFourYes{} & \TableFourYes{} & \TableFourNo{} & TXT/\allowbreak{}L/\allowbreak{}C/\allowbreak{}TAB/\allowbreak{}I & Page & Ans. Ret. \\
OpenDocVQA~\citeyearpar{ref11}~\MethodDate{25.04} & CVPR\allowbreak\textquoteright{}25 & Open & -- & 231.3k & -- & 43.5k & \TableFourYes{} & \TableFourYes{} & \TableFourYes{} & \TableFourNo{} & TXT/\allowbreak{}L/\allowbreak{}C/\allowbreak{}TAB/\allowbreak{}I & Page & Ans. Ret. \\
DocBench~\citeyearpar{ref12}~\MethodDate{24.07} & KNLP\allowbreak\textquoteright{}25\textsuperscript{W} & Closed & 229 & 66 & 46.4k & 1.1k & \TableFourNo{} & \TableFourYes{} & \TableFourYes{} & \TableFourYes{}(11.3\%) & TXT/\allowbreak{}L/\allowbreak{}C/\allowbreak{}TAB/\allowbreak{}I & -- & Ans. \\
Wiki-VISA; Paper-VISA~\citeyearpar{ref13}~\MethodDate{24.12} & ACL\allowbreak\textquoteright{}25 & Open & -- & -- & -- & 90k; 102.2k & \TableFourNo{} & \TableFourNo{} & \TableFourYes{} & \TableFourYes{} & TXT/\allowbreak{}TAB/\allowbreak{}I & Page, Elem. & Ans. Ret. Loc. \\
LongDocURL~\citeyearpar{ref14}~\MethodDate{24.12} & ACL\allowbreak\textquoteright{}25 & Closed & 396 & 85.6 & 43.6k & 2.3k & \TableFourNo{} & \TableFourYes{}(52.9\%) & \TableFourYes{}(47.0\%) & \TableFourYes{}(0.1\%) & TXT/\allowbreak{}L/\allowbreak{}C/\allowbreak{}TAB/\allowbreak{}I & Page & Ans. Ret. \\
AIRQA-REAL~\citeyearpar{ref156}~\MethodDate{25.05} & ACL\allowbreak\textquoteright{}25 & Open/\allowbreak{}Closed & 6.8k & -- & -- & 553 & \TableFourYes{}(27\%) & -- & -- & \TableFourNo{} & TXT/\allowbreak{}TAB/\allowbreak{}I & Doc. & Ans. Ret.\textsuperscript{a} \\
MMDocRAG*~\citeyearpar{ref105}~\MethodDate{25.05} & NIPS\allowbreak\textquoteright{}25\textsuperscript{D} & Closed & 222 & 67.0 & -- & 4.06k & \TableFourNo{} & \TableFourYes{}(52.0\%) & \TableFourYes{}(48.0\%) & \TableFourNo{} & TXT/\allowbreak{}C/\allowbreak{}TAB/\allowbreak{}I & Page, Elem. & Ans. Ret.\textsuperscript{a} \\
M-LongDoc~\citeyearpar{ref16}~\MethodDate{24.11} & EMNLP\allowbreak\textquoteright{}25 & Closed & 180 & 210.8 & 121.0k & 0.9k & \TableFourNo{} & \TableFourNo{} & \TableFourYes{} & \TableFourNo{} & TXT/\allowbreak{}C/\allowbreak{}TAB/\allowbreak{}I & Page & Ans. Ret. \\
ViDoSeek~\citeyearpar{ref18}~\MethodDate{25.02} & EMNLP\allowbreak\textquoteright{}25 & Open & 300 & 6.0k & -- & 1.14k & \TableFourNo{} & \TableFourNo{} & \TableFourYes{} & \TableFourNo{} & TXT/\allowbreak{}L/\allowbreak{}C/\allowbreak{}TAB/\allowbreak{}I & Page & Ans. Ret. \\
FinRAGBench-V~\citeyearpar{ref109}~\MethodDate{25.05} & EMNLP\allowbreak\textquoteright{}25 & Open & 2.21k & 57.1k & -- & 1.39k & -- & \TableFourYes{} & \TableFourYes{} & \TableFourNo{} & TXT/\allowbreak{}C/\allowbreak{}TAB & Page, Elem. & Ans. Ret. Loc. \\
M3DocVQA~\citeyearpar{cho2025m3docvqa}~\MethodDate{24.11} & ICCV\allowbreak\textquoteright{}25\textsuperscript{W} & Open & 3,368 & 41.0k & -- & 2.4k & \TableFourYes{} & \TableFourYes{} & \TableFourYes{} & \TableFourNo{} & TXT/\allowbreak{}TAB/\allowbreak{}I & Doc. & Ans. Ret.\textsuperscript{a} \\
Double-Bench~\citeyearpar{ref133}~\MethodDate{25.08} & AAAI\allowbreak\textquoteright{}26 & Open & 3.3k & 72.9k & -- & 5.2k & -- & \TableFourYes{} & \TableFourYes{} & \TableFourNo{} & TXT/\allowbreak{}L/\allowbreak{}C/\allowbreak{}TAB/\allowbreak{}I & Page & Ans. Ret. \\
AirQA~\citeyearpar{ref157}~\MethodDate{25.09} & ICLR\allowbreak\textquoteright{}26 & Open/\allowbreak{}Closed & 14.0k & -- & -- & 1.2k & \TableFourYes{} & -- & -- & \TableFourNo{} & TXT/\allowbreak{}TAB/\allowbreak{}I & Doc. & Ans. Ret.\textsuperscript{a} \\
M4DocBench~\citeyearpar{dong2026docresearcher}~\MethodDate{25.10} & WWW\allowbreak\textquoteright{}26 & Open & 303 & 306.4 & -- & 158 & \TableFourYes{}(91.1\%) & \TableFourYes{} & \TableFourYes{} & \TableFourNo{} & TXT/\allowbreak{}C/\allowbreak{}TAB/\allowbreak{}I & Page, Elem. & Ans. Ret.\textsuperscript{a} Loc. \\
DVisRAG~\citeyearpar{ref89}~\MethodDate{26.02} & CVPR\allowbreak\textquoteright{}26 & Open & -- & -- & -- & 367.6k & \TableFourNo{} & \TableFourYes{} & \TableFourYes{} & \TableFourNo{} & TXT/\allowbreak{}L/\allowbreak{}C/\allowbreak{}TAB/\allowbreak{}I & Page & Ans. Ret. Rob. \\
VinQA~\citeyearpar{ref98}~\MethodDate{26.06} & CVPR\allowbreak\textquoteright{}26 & Open & -- & 5.0 & -- & 44.3k & \TableFourYes{} & \TableFourYes{} & \TableFourYes{} & \TableFourYes{} & TXT/\allowbreak{}C/\allowbreak{}TAB/\allowbreak{}I & Page, Elem. & Ans. Ret. Loc. \\
ViDoRe V3~\citeyearpar{ref83}~\MethodDate{26.01} & ACL\allowbreak\textquoteright{}26 & Open & -- & -- & -- & 3.1k & \TableFourYes{} & \TableFourYes{} & \TableFourYes{} & -- & TXT/\allowbreak{}L/\allowbreak{}C/\allowbreak{}TAB/\allowbreak{}I & Page, Elem. & Ans. Ret. Loc. \\
PaperScope*~\citeyearpar{ref136}~\MethodDate{26.04} & ACL\allowbreak\textquoteright{}26\textsuperscript{F} & Open & 2.2k & -- & -- & 2.4k & \TableFourYes{}(91.7\%) & \TableFourYes{} & -- & \TableFourNo{} & TXT/\allowbreak{}L/\allowbreak{}C/\allowbreak{}TAB/\allowbreak{}I & Doc. & Ans. Ret.\textsuperscript{a} \\
M5BookVQA~\citeyearpar{ref57}~\MethodDate{26.07} & ACL\allowbreak\textquoteright{}26 & Open & -- & 16.8k & -- & 2.1k & -- & \TableFourYes{} & -- & -- & TXT/\allowbreak{}C/\allowbreak{}TAB/\allowbreak{}I & Page & Ans. Ret. \\
Wiki-CoE~\citeyearpar{ref111}~\MethodDate{26.05} & SIGIR\allowbreak\textquoteright{}26 & Open & 76.0k & 5.0 & -- & 70.4k & \TableFourYes{} & \TableFourYes{} & \TableFourNo{} & \TableFourNo{} & TXT/\allowbreak{}TAB/\allowbreak{}I/\allowbreak{}L & Page, Elem. & Ans. Ret. Loc. \\
MMLBD-C~\citeyearpar{ref74}~\MethodDate{26.02} & arXiv\allowbreak\textquoteright{}26 & Closed & 135 & 47.5 & 21.2k & 1.1k & \TableFourNo{} & \TableFourYes{} & \TableFourYes{} & \TableFourYes{} & TXT/\allowbreak{}L/\allowbreak{}C/\allowbreak{}TAB/\allowbreak{}I & Page & Ans. Ret. \\
MMDA~\citeyearpar{ref94}~\MethodDate{26.02} & PACMMOD\allowbreak\textquoteright{}26 & Closed & 537 & -- & -- & 1.1k & \TableFourNo{} & -- & -- & -- & TXT/\allowbreak{}L/\allowbreak{}C/\allowbreak{}TAB/\allowbreak{}I & -- & Ans. \\
SynthDocBench~\citeyearpar{ref71}~\MethodDate{26.07} & COLM\allowbreak\textquoteright{}26 & Closed & 200 & 51.1 & -- & 1.8k & \TableFourNo{} & \TableFourYes{} & \TableFourYes{} & \TableFourNo{} & TXT/\allowbreak{}C/\allowbreak{}TAB & Elem. & Ans. Ret.\textsuperscript{a} \\
DocVQA-2026~\citeyearpar{ref66}~\MethodDate{26.08} & ICDAR\allowbreak\textquoteright{}26\textsuperscript{C} & Closed & -- & -- & -- & 240 & \TableFourNo{} & \TableFourYes{} & \TableFourYes{} & \TableFourYes{} & TXT/\allowbreak{}L/\allowbreak{}C/\allowbreak{}TAB/\allowbreak{}I & -- & Ans. \\
LFDocQA~\citeyearpar{ref130}~\MethodDate{26.04} & arXiv\allowbreak\textquoteright{}26 & Open & 964 & -- & -- & 1.0k & \TableFourNo{} & \TableFourNo{} & \TableFourYes{} & \TableFourNo{} & TXT/\allowbreak{}L/\allowbreak{}C/\allowbreak{}TAB & Page, Elem. & Ans. Ret.\textsuperscript{a} Loc. \\
DocScope~\citeyearpar{ref113}~\MethodDate{26.05} & NIPS\allowbreak\textquoteright{}26\textsuperscript{D} & Closed & 273 & 51.3 & 24.6k & 1.12k & \TableFourNo{} & \TableFourYes{}(57.7\%) & \TableFourYes{}(35.3\%) & \TableFourYes{}(6.9\%) & TXT/\allowbreak{}C/\allowbreak{}TAB/\allowbreak{}I/\allowbreak{}L & Page, Elem. & Ans. Ret. Loc. \\
XL-DocBench~\citeyearpar{ref58}~\MethodDate{26.07} & NIPS\allowbreak\textquoteright{}26 & Open & 331 & -- & 227.5k & 1.5k & \TableFourYes{}(10.9\%) & \TableFourYes{}(72.6\%) & \TableFourYes{} & \TableFourYes{}(14.4\%) & TXT/\allowbreak{}C/\allowbreak{}TAB/\allowbreak{}I & Page & Ans. Ret. \\
\bottomrule
\end{tabular}
}
\endgroup
\par\vspace{1.5pt}\nointerlineskip
\begin{minipage}{\textwidth}
\fontsize{7.5}{8.5}\selectfont\raggedright
\textbf{1. Groups and dates.} Benchmarks are grouped by primary evaluation target and input scale. Ordering follows Table~\ref{tab:explicit-methods}. (YY.MM) denotes the first-release month, preferentially the first arXiv version; an earlier official release, or the official release when no preprint exists, takes precedence.\par\vspace{1pt}
\textbf{2. Venue.} Superscripts F/W/D/S/I/C denote Findings, Workshop, Datasets and Benchmarks, Short, Industry, and competition reports, respectively.\par\vspace{1pt}
\textbf{3. Scale and input.} Open and Closed denote a cross-document candidate collection and a given single document; Open/Closed supports both. \#Docs counts documents or independent page inputs. \#Pages is the average number of input pages per question, using a document-level mean when per-question statistics are unavailable; Open counts the full candidate collection for that question, and an independent image counts as one page. \#Tokens is the mean text-token count per input document. \#QA counts questions, queries, or localization tasks.\par\vspace{1pt}
\textbf{4. Question Type.} MD, MP, and SP denote multi-document, multi-page, and single-page evidence requirements, respectively; Unans. denotes questions unanswerable from the given evidence. \MethodYes{}/\MethodNo{} indicates inclusion/exclusion; parentheses give proportions. In Source, TXT/L/C/TAB/I denotes text/layout/charts/tables/images. Gran. denotes the verification granularity supported by public evidence annotations. In open-domain settings, page annotations with document identities also support document verification, which is not listed again.\par\vspace{1pt}
\textbf{5. Evaluation.} Eval. abbreviates Evaluation. Capabilities supported by the benchmark and its public annotations: Ans., answer quality; Ret., evidence retrieval, typically page/whole-image hits, recall, or ranking; Loc., within-page element or region localization, such as IoU-based evaluation; Eff., processing efficiency and cost; Rob., robustness to controlled perturbations. \textbf{Exceptions (a):} ViDoRe supports Ans. only on subsets with public answers. For Ret., SPIQA retrieves figures/tables; DocMMIR, DocViRe, PaperScope, AIRQA-REAL, and AirQA retrieve documents; M3DocVQA measures recall by the documents containing retrieved pages; M4DocBench covers documents, pages, and layout elements; SynthDocBench matches chart/element IDs; UDA measures text-evidence coverage; MMDocRAG includes text/image-segment selection; LFDocQA includes element-ID matching; PaperPDF matches paragraph/figure/table IDs. Loc. for GroundingDocQA-Bench and DoCoG-Bench includes segmentation masks.\par\vspace{1pt}
\textbf{6. Pre-extracted inputs (*).} ArXivQA uses pre-extracted scientific figures; SPIQA uses pre-extracted figure/table images and text; MMDocRAG generation evaluation uses pre-extracted text/image segments; PaperScope supplies pre-parsed paper text and images. These settings do not directly evaluate the complete parsing process from raw pages.\par\vspace{1pt}
\end{minipage}
\end{table*}

\clearpage
\endgroup

\begingroup
\newcommand{\TableFiveYes}{\MethodYes}
\newcommand{\TableFiveNo}{\MethodNo}
\makeatletter
\setlength{\@fptop}{0pt}
\setlength{\@dblfptop}{0pt}
\makeatother

\begin{table*}[!t]
\captionsetup{position=top,skip=3pt}
\caption{Complete comparison of 29 methods without explicit retrieval, grouped by their primary single-page or multipage setting. Ordering and first-release dates follow Table~\ref{tab:explicit-methods}. + denotes combinations and / alternatives; -- denotes unconfirmed information.}
\label{tab:direct-methods-full}
\centering
\begingroup
\fontsize{7.1}{8.0}\selectfont
\setlength{\tabcolsep}{1.15pt}
\renewcommand{\arraystretch}{0.96}
\setlength{\aboverulesep}{1.0pt}
\setlength{\belowrulesep}{1.0pt}
\newsavebox{\TableFiveBox}
\sbox{\TableFiveBox}{%
\begin{tabular}{@{}lcccclll@{}}
\toprule
\multicolumn{2}{c}{\textbf{Method}} &
\multicolumn{2}{c}{\textbf{Parsing}} &
\multicolumn{3}{c}{\textbf{Generation \& Learning}} &
\multirow{2}{*}{\textbf{Eval.}} \\
\cmidrule(lr){1-2}\cmidrule(lr){3-4}\cmidrule(lr){5-7}
\textbf{Name} & \textbf{Venue} & \textbf{OCR} & \textbf{Deep} & \textbf{Agent} & \textbf{Model} & \textbf{Learning} & \\
\midrule
\multicolumn{8}{@{}l@{}}{\textit{Single-page VQA}} \\
\addlinespace[1.0pt]
VisionTaPas~\citeyearpar{ref75}~\MethodDate{22.03} & ACL\allowbreak\textquoteright{}22\textsuperscript{F} & \TableFiveYes{} & \TableFiveNo{} & Single & TAPAS & SFT & Ans. \\
Donut~\citeyearpar{ref67}~\MethodDate{21.11} & ECCV\allowbreak\textquoteright{}22 & \TableFiveNo{} & \TableFiveNo{} & Single & BART & Pretraining + SFT & Ans. Eff. \\
LayoutLMv3~\citeyearpar{huang2022layoutlmv3}~\MethodDate{22.04} & MM\allowbreak\textquoteright{}22 & \TableFiveYes{} & \TableFiveNo{} & Single & RoBERTa & Pretraining + SFT & Ans. Loc.\textsuperscript{a} \\
UDOP~\citeyearpar{ref87}~\MethodDate{22.12} & CVPR\allowbreak\textquoteright{}23 & \TableFiveYes{} & \TableFiveNo{} & Single & T5 & Pretraining + SFT & Ans. \\
Pix2Struct~\citeyearpar{ref76}~\MethodDate{22.10} & ICML\allowbreak\textquoteright{}23 & \TableFiveNo{} & \TableFiveNo{} & Single & Pix2Struct & Pretraining + SFT & Ans. \\
UReader~\citeyearpar{ref92}~\MethodDate{23.10} & EMNLP\allowbreak\textquoteright{}23\textsuperscript{F} & \TableFiveNo{} & \TableFiveNo{} & Single & Llama & SFT & Ans. Eff. \\
LayoutLLM (Fujitake)~\citeyearpar{ref124}~\MethodDate{24.03} & LREC-COLING\allowbreak\textquoteright{}24 & \TableFiveYes{} & \TableFiveNo{} & Single & Llama & SFT & Ans. \\
mPLUG-DocOwl 1.5~\citeyearpar{ref93}~\MethodDate{24.03} & EMNLP\allowbreak\textquoteright{}24\textsuperscript{F} & \TableFiveNo{} & \TableFiveNo{} & Single & Llama & Pretraining + SFT & Ans. Eff. \\
DocLayLLM~\citeyearpar{ref22}~\MethodDate{24.08} & CVPR\allowbreak\textquoteright{}25 & \TableFiveYes{} & \TableFiveNo{} & Single & Llama & Pretraining + SFT & Ans. Eff. \\
DocVLM*~\citeyearpar{ref25}~\MethodDate{24.12} & CVPR\allowbreak\textquoteright{}25 & \TableFiveYes{} & \TableFiveNo{} & Single & Qwen-VL /\allowbreak{} InternVL /\allowbreak{} LLaVA & SFT & Ans. Eff. \\
DOGR~\citeyearpar{ref15}~\MethodDate{24.11} & ICCV\allowbreak\textquoteright{}25 & \TableFiveNo{} & \TableFiveNo{} & Single & Qwen & Pretraining + SFT & Ans. Loc. \\
M3Grounder~\citeyearpar{ref119}~\MethodDate{26.06} & CVPR\allowbreak\textquoteright{}26 & \TableFiveNo{} & \TableFiveNo{} & Single & InternVL /\allowbreak{} Qwen-VL & SFT & Ans. Loc.\textsuperscript{a} \\
DocVAL~\citeyearpar{ref23}~\MethodDate{25.11} & ICML\allowbreak\textquoteright{}26 & \TableFiveNo{} & \TableFiveNo{} & Single & Gemma3 & SFT & Ans. Loc. Eff. \\
Uni-DocRobust~\citeyearpar{ref24}~\MethodDate{26.-{}-} & ICML\allowbreak\textquoteright{}26 & \TableFiveNo{} & \TableFiveNo{} & Single & LLaVA /\allowbreak{} Qwen-VL /\allowbreak{} InternVL & Pretraining & Ans. Eff. Rob. \\
DoCoG~\citeyearpar{ref120}~\MethodDate{26.09} & ECCV\allowbreak\textquoteright{}26 & \TableFiveNo{} & \TableFiveNo{} & Single & Qwen-VL & SFT + RL & Ans. Loc.\textsuperscript{a} \\
RegionDoc-R1*~\citeyearpar{ref140}~\MethodDate{26.-{}-} & EMNLP\allowbreak\textquoteright{}26 & \TableFiveNo{} & \TableFiveNo{} & Single & Qwen-VL & SFT + RL & Ans. \\
\midrule
\multicolumn{8}{@{}l@{}}{\textit{Multi-page VQA}} \\
\addlinespace[1.0pt]
M3D~\citeyearpar{tanaka2023slidevqa}~\MethodDate{23.01} & AAAI\allowbreak\textquoteright{}23 & \TableFiveYes{} & \TableFiveNo{} & Single & T5 & SFT & Ans. Ret. \\
Hi-VT5~\citeyearpar{tito2023hierarchical}~\MethodDate{22.12} & PR\allowbreak\textquoteright{}23 & \TableFiveYes{} & \TableFiveNo{} & Single & T5 & Pretraining + SFT & Ans. Ret. \\
GRAM~\citeyearpar{ref117}~\MethodDate{24.01} & CVPR\allowbreak\textquoteright{}24 & \TableFiveYes{} & \TableFiveNo{} & Single & T5 & SFT & Ans. Eff. \\
Docopilot~\citeyearpar{ref139}~\MethodDate{25.06} & CVPR\allowbreak\textquoteright{}25 & \TableFiveNo{} & \TableFiveNo{} & Single & InternVL & SFT & Ans. Eff. \\
mPLUG-DocOwl2~\citeyearpar{ref26}~\MethodDate{24.09} & ACL\allowbreak\textquoteright{}25 & \TableFiveNo{} & \TableFiveNo{} & Single & Llama & Pretraining + SFT & Ans. Eff. \\
DocR1~\citeyearpar{ref27}~\MethodDate{25.08} & AAAI\allowbreak\textquoteright{}26 & \TableFiveNo{} & \TableFiveNo{} & Single & Qwen-VL & RL & Ans. Ret. \\
LAT~\citeyearpar{ref110}~\MethodDate{25.11} & AAAI\allowbreak\textquoteright{}26 & \TableFiveNo{} & \TableFiveNo{} & Single & Qwen-VL & SFT + RL & Ans. Ret. Loc. Eff. \\
DocSLM~\citeyearpar{ref63}~\MethodDate{25.11} & CVPR\allowbreak\textquoteright{}26\textsuperscript{F} & \TableFiveYes{} & \TableFiveNo{} & Single & Qwen & Pretraining + SFT & Ans. Eff. \\
DocSeeker~\citeyearpar{ref29}~\MethodDate{26.04} & CVPR\allowbreak\textquoteright{}26 & \TableFiveNo{} & \TableFiveNo{} & Single & Qwen-VL & SFT + RL & Ans. Ret. Eff. Rob. \\
VinQA (Page Encoding)~\citeyearpar{ref98}~\MethodDate{26.06} & CVPR\allowbreak\textquoteright{}26 & \TableFiveNo{} & \TableFiveNo{} & Single & Qwen-VL & SFT & Ans. Ret.\textsuperscript{a} Eff. \\
CoE~\citeyearpar{ref111}~\MethodDate{26.05} & SIGIR\allowbreak\textquoteright{}26 & \TableFiveNo{} & \TableFiveNo{} & Single & Qwen-VL & SFT & Ans. Ret. Loc.\textsuperscript{a} Eff. \\
OriOn~\citeyearpar{ref74}~\MethodDate{26.02} & arXiv\allowbreak\textquoteright{}26 & \TableFiveNo{} & \TableFiveNo{} & Single & Mistral /\allowbreak{} Qwen-VL & Pretraining + SFT + RL & Ans. Eff. \\
Quote-and-Retrieve~\citeyearpar{ref112}~\MethodDate{26.07} & arXiv\allowbreak\textquoteright{}26 & \TableFiveNo{} & \TableFiveNo{} & Single & Qwen-VL & RL & Ans. Ret. Loc. Eff. \\
\bottomrule
\end{tabular}%
}
\ifdim\wd\TableFiveBox>\textwidth
  \resizebox{\textwidth}{!}{\usebox{\TableFiveBox}}%
\else
  \usebox{\TableFiveBox}%
\fi
\endgroup
\par\vspace{1.5pt}\nointerlineskip
\begin{minipage}{\textwidth}
\fontsize{7.5}{8.5}\selectfont\raggedright
\textbf{1. Methods.} * marks methods primarily evaluated on single pages/images that also support multiple pages. Venue superscripts follow Table~\ref{tab:benchmark-catalog}; (YY.MM) denotes first release, and -- in the month position means unavailable. Deep = Deep Parsing; Eval. = Evaluation.\par\vspace{1pt}
\textbf{2. Model and learning.} Model gives the main answer model or language-backbone family. Learning records added Pretraining, supervised fine-tuning (SFT), or reinforcement learning (RL).\par\vspace{1pt}
\textbf{3. Evaluation.} Ans./Ret./Loc./Eff./Rob. denote answer quality/retrieval/localization/efficiency/robustness. Only dimensions quantitatively reported in the original paper are listed. Ret. denotes page-level evidence selection or prediction and does not imply explicit retrieval in the answering pipeline. \textbf{Exceptions (a):} LayoutLMv3 Loc. is layout-box detection; VinQA (Page Encoding) Ret. is element-citation F1; CoE Loc. includes box-center hits; M3Grounder and DoCoG Loc. include segmentation masks.\par\vspace{1pt}
\end{minipage}
\end{table*}

\clearpage
\endgroup

\onecolumn
\begingroup
\ifdefined\nolinenumbers\nolinenumbers\fi
\fontsize{6.3}{7.2}\selectfont
\setlength{\tabcolsep}{0.65pt}
\renewcommand{\arraystretch}{1.0}
\setlength{\extrarowheight}{0pt}
\setlength{\aboverulesep}{1.0pt}
\setlength{\belowrulesep}{1.0pt}
\setlength{\LTleft}{0pt plus 1fill}
\setlength{\LTright}{0pt plus 1fill}
\setlength{\LTcapwidth}{\textwidth}
\setlength{\LTpre}{0pt}
\setlength{\LTpost}{5pt}
\setlength{\emergencystretch}{0.6em}
\renewcommand{\MethodDate}[1]{{\fontsize{5.2}{6.0}\selectfont\textcolor{MethodDateColor}{\textit{(#1)}}}}

\newsavebox{\TableSixCellBox}
\newcommand{\TableSixFit}[1]{%
  \begingroup
  \sbox{\TableSixCellBox}{#1}%
  \ifdim\wd\TableSixCellBox>\linewidth
    \ifdim\wd\TableSixCellBox>1.12\linewidth
      #1%
    \else
      \leavevmode\resizebox{\linewidth}{!}{\usebox{\TableSixCellBox}}%
    \fi
  \else
    #1%
  \fi
  \endgroup
}

\begin{longtable}{@{}>{\raggedright\arraybackslash}p{\dimexpr0.164\textwidth-2\tabcolsep\relax}>{\centering\arraybackslash}p{\dimexpr0.075\textwidth-2\tabcolsep\relax}>{\centering\arraybackslash}p{\dimexpr0.034\textwidth-2\tabcolsep\relax}>{\centering\arraybackslash}p{\dimexpr0.064\textwidth-2\tabcolsep\relax}>{\centering\arraybackslash}p{\dimexpr0.071\textwidth-2\tabcolsep\relax}>{\centering\arraybackslash}p{\dimexpr0.065\textwidth-2\tabcolsep\relax}>{\raggedright\arraybackslash}p{\dimexpr0.160\textwidth-2\tabcolsep\relax}>{\centering\arraybackslash}p{\dimexpr0.058\textwidth-2\tabcolsep\relax}>{\centering\arraybackslash}p{\dimexpr0.043\textwidth-2\tabcolsep\relax}>{\centering\arraybackslash}p{\dimexpr0.067\textwidth-2\tabcolsep\relax}>{\raggedright\arraybackslash}p{\dimexpr0.069\textwidth-2\tabcolsep\relax}>{\centering\arraybackslash}p{\dimexpr0.058\textwidth-2\tabcolsep\relax}>{\raggedright\arraybackslash}p{\dimexpr0.072\textwidth-2\tabcolsep\relax}@{}}
\caption{Complete comparison of methods with explicit retrieval: 20 retrieval-only and 90 document VQA methods. Ordering and date conventions follow Table~\ref{tab:explicit-methods}; venue superscripts follow Table~\ref{tab:benchmark-catalog}. Mechanism specifies retrievers, evidence mappings, and retrieval training; Model gives the main generator model family. + denotes combinations and / alternatives; -- denotes unreported, unconfirmed, or inapplicable information.}
\label{tab:explicit-methods-full}\\
\toprule
\multicolumn{2}{c}{\textbf{Method}} &
\multicolumn{2}{c}{\textbf{Parsing}} &
\multicolumn{4}{c}{\textbf{Evidence Organization and Retrieval}} &
\multicolumn{4}{c}{\textbf{Generation \& Learning}} &
\multicolumn{1}{c}{\multirow{2}{*}{\textbf{Eval.}}} \\
\cmidrule(lr){1-2}\cmidrule(lr){3-4}\cmidrule(lr){5-8}\cmidrule(lr){9-12}
\mbox{\textbf{Name}} & \mbox{\textbf{Venue}} & \mbox{\textbf{OCR}} & \mbox{\textbf{Deep}} & \mbox{\textbf{Modality}} & \mbox{\textbf{Gran.}} & \mbox{\textbf{Mechanism}} & \mbox{\textbf{Acq.}} & \mbox{\textbf{Agent}} & \mbox{\textbf{Reasoning}} & \mbox{\textbf{Model}} & \mbox{\textbf{Learning}} & \\
\midrule
\multicolumn{13}{@{}l@{}}{\textit{Retrieval-only methods}} \\*
\endfirsthead
\multicolumn{13}{@{}l@{}}{\textbf{Table \thetable\ (continued)}} \\
\toprule
\multicolumn{2}{c}{\textbf{Method}} &
\multicolumn{2}{c}{\textbf{Parsing}} &
\multicolumn{4}{c}{\textbf{Evidence Organization and Retrieval}} &
\multicolumn{4}{c}{\textbf{Generation \& Learning}} &
\multicolumn{1}{c}{\multirow{2}{*}{\textbf{Eval.}}} \\
\cmidrule(lr){1-2}\cmidrule(lr){3-4}\cmidrule(lr){5-8}\cmidrule(lr){9-12}
\mbox{\textbf{Name}} & \mbox{\textbf{Venue}} & \mbox{\textbf{OCR}} & \mbox{\textbf{Deep}} & \mbox{\textbf{Modality}} & \mbox{\textbf{Gran.}} & \mbox{\textbf{Mechanism}} & \mbox{\textbf{Acq.}} & \mbox{\textbf{Agent}} & \mbox{\textbf{Reasoning}} & \mbox{\textbf{Model}} & \mbox{\textbf{Learning}} & \\
\midrule
\endhead
\midrule
\multicolumn{13}{r@{}}{\textit{Continued on next page}} \\
\endfoot
\bottomrule
\multicolumn{13}{r@{}}{\textit{End of Table \thetable}} \\
\endlastfoot
DSE {\fontsize{5.3}{6.1}\selectfont \citeyearpar{ma2024unifying} \MethodDate{24.06}} & \TableSixFit{EMNLP\allowbreak\textquoteright{}24} & \MethodNo{} & \MethodNo{} & Uni\allowbreak{}(V)\allowbreak{} & \TableSixFit{Page} & \TableSixFit{DSE\textsuperscript{t}\allowbreak{}(P\textsuperscript{V})\allowbreak{}} & \mbox{One-shot} & -- & -- & -- & -- & \TableSixFit{Ret. Eff.} \\
ColPali {\fontsize{5.3}{6.1}\selectfont \citeyearpar{ref7} \MethodDate{24.06}} & \TableSixFit{ICLR\allowbreak\textquoteright{}25} & \MethodNo{} & \MethodNo{} & Uni\allowbreak{}(V)\allowbreak{} & \TableSixFit{Page} & \TableSixFit{ColPali\textsuperscript{t}\allowbreak{}(P\textsuperscript{V})\allowbreak{}} & \mbox{One-shot} & -- & -- & -- & -- & \TableSixFit{Ret. Eff.} \\
RobTabCol {\fontsize{5.3}{6.1}\selectfont \citeyearpar{ref88} \MethodDate{25.02}} & \TableSixFit{ACL\allowbreak\textquoteright{}25} & \MethodNo{} & \MethodNo{} & Uni\allowbreak{}(V)\allowbreak{} & \TableSixFit{Page} & \TableSixFit{RobTabCol\textsuperscript{t}\allowbreak{}(P\textsuperscript{V})\allowbreak{}} & \mbox{One-shot} & -- & -- & -- & -- & \TableSixFit{Ret. Rob.} \\
Light-ColPali {\fontsize{5.3}{6.1}\selectfont \citeyearpar{ref31} \MethodDate{25.06}} & \TableSixFit{ACL\allowbreak\textquoteright{}25\textsuperscript{F}} & \MethodNo{} & \MethodNo{} & Uni\allowbreak{}(V)\allowbreak{} & \TableSixFit{Page} & \TableSixFit{ColPali\textsuperscript{t}\allowbreak{}(P\textsuperscript{V})\allowbreak{}} & \mbox{One-shot} & -- & -- & -- & -- & \TableSixFit{Ret. Eff.} \\
DocMMIR {\fontsize{5.3}{6.1}\selectfont \citeyearpar{ref19} \MethodDate{25.05}} & \TableSixFit{EMNLP\allowbreak\textquoteright{}25\textsuperscript{F}} & \MethodYes{} & \MethodNo{} & \mbox{Mixed(T/V)} & \TableSixFit{Doc.} & \TableSixFit{OpenCLIP\textsuperscript{t}\allowbreak{}(D\textsuperscript{TV})\allowbreak{}} & \mbox{One-shot} & -- & -- & -- & -- & \TableSixFit{Ret.\textsuperscript{a} Eff.} \\
\end{longtable}
\ifdefined\nolinenumbers\nolinenumbers\fi
\par\vspace{1.5pt}\nointerlineskip
\noindent\begin{minipage}{\textwidth}
\ifdefined\nolinenumbers\nolinenumbers\fi
\fontsize{7.5}{8.5}\selectfont
\raggedright
\textbf{Header abbreviations.} Deep = Deep Parsing; Gran. = Granularity; Acq. = Acquisition; Eval. = Evaluation.\par\vspace{1pt}
\textbf{Type labels.} Deep and Learning list structure and training types directly; a standalone \MethodYes{} in Learning denotes added training without a listed type. SFT/RL: supervised fine-tuning/reinforcement learning.\par\vspace{1pt}
\textbf{1. Mechanism.} D/P/E denotes document/page/element; superscripts T/V/TV denote text/visual/both. The left side of $\rightarrow$ is the evidence representation returned by retrieval or selection; the right side is the generation input. Without an arrow, retrieval and generation use the same granularity and modality; retrieval-only methods are labeled by their returned unit. Superscript t denotes additional retrieval, scoring, or page-selection training. Parentheses after model names specify evidence forms; semicolons separate independent paths, and an arrow applies only within its path. Brackets [ ] group multiple evidence items, with commas denoting parallel items.\par\vspace{1pt}
\textbf{2. Evaluation.} Only dimensions quantitatively reported in the original paper are listed. Ret. evaluates evidence retrieval, usually page-level hits, recall, or ranking; Loc. evaluates within-page element/region localization or coverage, such as IoU or region-overlap recall. \textbf{Exceptions (a):} Ret. is document retrieval for DocMMIR, DocPC, NeuSym-RAG, and Agentic Hybrid (EXTRACTOR), document identification for HiKEY, and additionally document recall for HEAVEN; LILaC evaluates layout-element retrieval, and Doc-Researcher evaluates document/page/layout-element retrieval. G\textsuperscript{2}-Reader Ret. measures semantic coverage of evidence statements; LFRAG includes element-ID matching; HierDoc includes region-ID-set F1; MoDora evaluates evidence-node recall and filtering. ReAlign Loc. measures overlap between attention regions and supervised boxes; RGenCite includes VLM-assessed evidence support; Lang2Act measures the proportion of internal attention falling within gold regions. MMRetHeads evaluates pages and layout regions on MMDocIR; LMS-Retrieval Ret./Loc. measure layout-element recall and region coverage.\par\vspace{1pt}
\end{minipage}
\par
\endgroup

\end{document}